\documentclass{article}
\pdfoutput=1
\usepackage{caption}
\usepackage{subcaption}
\usepackage{graphicx,stfloats}
\usepackage{pdflscape}
\usepackage{rotating}
\usepackage{multirow}
\usepackage{epstopdf}
\usepackage{amsmath}
\usepackage{authblk}
\usepackage[T1]{fontenc}

\begin{document}

\renewcommand\Affilfont{\fontsize{9}{10.8}\itshape}

\title{\vspace{-2cm}A New Distance Measure for Non-Identical Data with Application to Image Classification}
\date{}

\author[1,2]{Muthukaruppan Swaminathan}
\author[1]{Pankaj Kumar Yadav}
\author[3]{Obdulio Piloto}
\author[4]{Tobias Sj\"oblom}
\author[1,2]{Ian Cheong\thanks{Corresponding author: ian@tll.org.sg}}
\affil[1]{Temasek Life Sciences Laboratory, Singapore}
\affil[2]{Dept of Biological Sciences, National University of Singapore}
\affil[3]{Entopsis LLC, USA}
\affil[3]{Rudbeck Laboratory, Uppsala University, Sweden}

\maketitle

\begin{abstract}
Distance measures are part and parcel of many computer vision algorithms. The underlying assumption in all existing distance measures is that feature elements are independent and identically distributed. However, in real-world settings, data generally originate from heterogeneous sources even if they do possess a common data-generating mechanism. Since these sources are not identically distributed by necessity, the assumption of identical distribution is inappropriate. Here, we use statistical analysis to show that feature elements of local image descriptors are indeed non-identically distributed. To test the effect of omitting the unified distribution assumption, we created a new distance measure called the Poisson-Binomial Radius (PBR). PBR is a bin-to-bin distance which accounts for the dispersion of bin-to-bin information. PBR's performance was evaluated on twelve benchmark data sets covering six different classification and recognition applications: texture, material, leaf, scene, ear biometrics and category-level image classification. Results from these experiments demonstrate that PBR outperforms state-of-the-art distance measures for most of the data sets and achieves comparable performance on the rest, suggesting that accounting for different distributions in distance measures can improve performance in classification and recognition tasks.
\end{abstract}


\section{Introduction}

 The basis for many computer vision algorithms is quantifying the pairwise similarity of feature vectors. This step is often accomplished by evaluating the distance between feature vectors using a distance (dissimilarity) measure. Thus, choosing an appropriate distance measure is fundamentally important as it determines the performance of learning algorithms in numerous applications such as image retrieval~\cite{Jacobs2000}, stereo matching~\cite{Sebe2000} and image classification~\cite{zhang2007},\cite{chapelle1999} to name a few.

Mathematically, a distance measure, $D(X,Y)$, between pairs of feature vectors, is considered a distance metric if the following axioms are satisfied:

\begin{enumerate}
	\item $D(X,Y)$ $\geq$ 0 (non-negativity)
	\item $D(X,Y)$ = 0 $\iff$ $X$ = $Y$ (identity of indiscernibles)
	\item $D(X,Y)$ = $D(Y,X)$ (symmetry)
	\item $D(X,Y)$ $\leq$ $D(Y,Z)$ + $D(X,Z)$ (triangle inequality)
	
\end{enumerate}

Despite the common use of distance metrics, Scheirer et al.\cite{scheirer2014} noted that top scoring computer vision algorithms tended to employ distance measures which are non-metric, meaning that they do not obey at least one of the four axioms. The authors of \cite{scheirer2014} state: ``What makes non-metric algorithms better? We emphasize that treating all samples alike may unnecessarily handicap an algorithm.'' The triangle inequality axiom deserves particular attention. A distance measure which does not obey the triangle inequality is by definition a semimetric distance. The consensus in the field of cognitive psychology is that human similarity judgments violate the triangle inequality. Tversky et al.~\cite{Tversky1982} specifically demonstrated that human subjects given visual pair-matching tasks violated the triangle inequality in a statistically significant manner. Given that human judgment is semimetric, we propose that the triangle inequality is at best unnecessary and at worst a constraint on classification performance.

Moreover, at the heart of all existing distance measures lies the assumption that data are independent and identically distributed (i.i.d). To restate the issue, all existing distance measures, regardless of whether they are parametric or non-parametric, assume that data derives from a single source. Hence, methods to calculate these distances must make the i.i.d. assumption. This may be exemplified at a basic level by considering non-parametric entropy-based distance measures such as Kullback-Liebler (KL) divergence and Jeffrey divergence (JD), which is the symmetric version of KL. Because measures of entropy are predicated on data being from a single source, all entropy-based measures such as KL and JD must by definition make the i.i.d. assumption when measuring information content. Similar arguments can be applied to all existing histogram distances.

Whether the i.i.d assumption is justified in real world settings is an often debated topic. Feature elements are obtained from heterogeneous sources, even if they are derived from a common data generating mechanism. Hence, it is reasonable to assume that these feature elements are not identically distributed~\cite{Tillman2009}. For example, image acquisition data derives from digitized imaging sensor outputs, noise caused by color filter array defects and pattern noise~\cite{khanna2006}. Given these heterogeneous sources, it is preferable from the viewpoint of parsimony to assume that feature elements originate from different distributions and are hence independent but non-identically distributed (i.n.i.d). For example, Yu et al.~\cite{yu2008} found that the best accuracy was achieved when the distribution for each feature element was independently modeled. In that study, single isotropic distributions could not be used to estimate the data distribution, hence corroborating the point that feature distributions in real data should be assumed to be heterogeneous. Such an approach however, requires \emph{a priori} knowledge of how feature elements are distributed. Also, one is limited by the set of distributions chosen to model the data distribution. In theory, a more generalizable solution would be to adopt a distance measure which avoids the assumption in the first instance. However, a distance measure for non-identically distributed data has hitherto not been proposed and tested. 

In this paper, we address the above-mentioned deficiency by proposing a new i.n.i.d. distance measure called the Poisson-Binomial Radius (PBR). The underlying basis for PBR is the Poisson-Binomial distribution which is the probability mass function of independent Bernoulli trials which are not necessarily identically-distributed. PBR is hence the first distribution-agnostic distance measure.

Our contributions are as follows:

\begin{enumerate}
	\item We show that feature elements of a local descriptor are indeed from significantly different distributions.
	\item We present PBR, a semimetric distance measure which avoids the identical distribution assumption and accounts for non-identically distributed feature elements.
	\item We evaluate PBR on six different image classification / recognition applications using twelve benchmark data sets and demonstrate its effective performance.
\end{enumerate}

The rest of the paper is organized as follows. In section 2, we review recent advances in image classification and commonly used distance measures. In section 3, we discuss the Poisson-Binomial distribution. The PBR semimetric distance is introduced in section 4. In section 5, we present our results from experimental comparisons between PBR and other distance measures using a kernel-based image classification framework. Concluding remarks are provided in section 6.

\section{Related work} 
\subsection{Image classification}

Classifying images according to their visual content still remains one of the most challenging problems in computer vision. This is largely due to the fact that real world images suffer from changes in viewpoint (e.g., translation, rotation, and scale), illumination variations, background clutter, partial occlusion, intra-class diversity and inter-class similarity. 

To efficiently address these issues, various hand crafted features (e.g., gradients, patterns, shapes and gray-tones/colors) have been presented in the literature (see ~\cite{li2015} for an overview). One interesting development is the integration of multiple such visual feature representations. The challenge for this learning framework, called multiview learning, lies in integrating features from different spatial viewpoints or multiple features from a single viewpoint. Luo et al.~\cite{luo2015} recently proposed a framework to fuse different feature descriptors for semisupervised multilabel classification using matrix completion. These matrix completion outputs were then combined using weights obtained through cross-validation on the labeled set. One consequence of multiview features is that some views are more important than others. Zhu et al.~\cite{zhu2016} addresses this problem using a block-row Frobenius norm regularizer-based framework to select informative views and features while avoiding redundant views and noisy features. This method outperformed baseline algorithms with low computational overhead. 

The rise in popularity of deep learning models has seemingly rendered hand crafted feature engineering otiose since these models learn a hierarchical feature representation directly from the image. It is however perhaps more appropriate to view deep learning as having replaced feature engineering with architecture engineering. The convolutional neural network (CNN) in particular is a deep learning architecture which has shown its promise for image classification tasks~\cite{krizhevsky2012}. This success has been driven by optimized implementations of algorithms for graphics processing unit (GPU) architectures and new large scale data sets such as LabelMe~\cite{russell2008} and ImageNet~\cite{ILSVRC15}. Moreover, improvements have been made to CNN classification performance and training time. He et al.~\cite{he2015} presented a residual learning framework in which inputs are added to the output of the stacked layers, thus asymptotically approximating the desired underlying mapping function. This framework is shown to be easily optimized and exhibits accuracy gain when compared to CNNs which are simply stacked. Clevert et al.~\cite{clevert2015} proposed the exponential linear unit (ELU) as a replacement for the rectified linear unit (ReLU), currently the most popular activating function in CNNs. ELUs enable faster learning because they saturate to negative values with smaller arguments. Besides pushing mean unit activations closer to zero, this behavior also helps CNNs to learn representations which are robust to noise. 

Recent works have also incorporated non-semantic features as inputs for CNNs. For example, Wang et al.~\cite{wang2016} used CNNs in combination with recurrent neural networks (RNNs) to perform multi-label image classification by learning a joint embedding space to characterize the image-label relationship and label dependency. In a similar vein, Tang et al.~\cite{tang2015} encoded GPS coordinates in a CNN model to classify images with location context.

Importantly, there has been a great deal of interest in hybridizing support vector machine (SVM) architectures with CNNs. In this hybrid model, CNNs work as an automatic feature extractor while decision surfaces are generated by SVMs. Compared to CNNs alone, CNN-SVM hybrids with non-linear kernels such as Gaussian and radial basis functions have been shown to achieve superior classification accuracy with benchmarking data sets~\cite{niu2012},\cite{huang2006},\cite{kim2015}. 

Even though CNNs are neither the focus nor compared in this work, the non-linear kernel proposed in this work can be effectively combined with CNNs in the manner described.

\subsection{Distance measures}
Several distance measures use histogram-based representations because of their simplicity and rich discriminative information. These measures may be broadly categorized as bin-to-bin, cross-bin or intra-cross-bin distances. 

Bin-to-bin distances compare corresponding bin values between two histograms. Summing these `same-bin' comparisons gives the resultant distance between two histograms. The most commonly used bin-to-bin distances are $L_1$ and $L_2$. Both belong to the Minkowski family and are distance metrics. Other bin-to-bin distance measures such as Bhattacharyya distance (BD)~\cite{you2010}, Jeffrey divergence (JD)~\cite{nguyen2010}, $\chi^2$ distance~\cite{guo2013},\cite{Vempati2010}, Hellinger distance~\cite{tran2011} and histogram intersection (HI)~\cite{vedaldi2012} are also widely used in computer vision applications. It is to be noted that histogram intersection is equivalent to $L_1$ distance when the area of the two histograms are equal. While BD, JD and $\chi^2$ distances are semimetric in nature, Hellinger distance and HI are distance metrics.

In contrast to bin-to-bin distances, cross-bin distances compare values between different bins and aggregate the distances using the ground distance. A notable example is the Earth Mover Distance (EMD)~\cite{Rubner2000} which is based on a solution to the well-known \emph{transportation problem}. EMD can be categorized as a distance metric if the ground distance is a metric. EMD has been shown to be very effective for comparing generic data summaries called \emph{signatures} ~\cite{zhang2007}.  However the computational complexity of EMD is $O(n^3 \log{}n)$ for a histogram with $n$ bins. On the other hand, match distance~\cite{werman1985}, a special form of EMD for one dimensional histograms with equal areas, is equivalent to the $L_1$ distance between the cumulative histograms. But this distance is not robust to partial matches. Another commonly used cross-bin distance measure is Mahalanobis distance~\cite{haasdonk2010}. This distance represents the relationship among the bins using the covariance matrix. Though it is scale-invariant, the covariance matrix is usually singular when the number of feature elements in the feature vectors is large. Further, when the feature vectors are sparse, the covariance matrix might not capture the distribution of the features~\cite{hu2014}.

The last category of distance measures is the intra-cross-bin distance. Hu et al.~\cite{hu2014} introduced this new category and called their prototype intra-cross-bin measure the Bin-Ratio-based histogram Distance (BRD). This distance operates on the ratios between histogram bin values, hence capturing correlations between them. BRD can be combined with $\chi^2$ and $L_1$ distances to form $\chi^2$-BRD and $L_1$-BRD respectively. These hybrid measures satisfy the triangle inequality axiom and hence qualify as distance metrics.

\section{Poisson-Binomial distribution}
\textbf{Definition.} \textit{The Poisson-Binomial distribution is defined by the probability mass function (p.m.f.) for $n$ successes given independent but non-identical probabilities $(p_1,\ldots,p_N)$ of success. The distribution is unimodal with mean $\mu = \sum_{i=1}^{N} p_i$ and variance $\sigma^2 = \sum_{i=1}^N (1-p_i) p_i$. If $\Omega_{n}$ is the corresponding sample space of all possible $\binom{N}{n}$ paired sets of $I$ and $I^C$ resulting from $n$ occurrences and $N-n$ non-occurrences, then the Poisson-Binomial p.m.f. is given as}

\begin{equation}
P(\Omega_n)= \sum_{j=1}^{\binom{N}{n}}\; {\prod_{i\in I_{j}}{p_i}\prod_{i\in I^{C}_{j}}{(1-p_i)}} 
\end{equation}

A special case of this distribution is the binomial distribution where $p_i$ has the same value for all $i$. The Poisson-Binomial distribution finds use in a broad range of fields such as biology, imaging, data mining, bioinformatics and engineering.

\indent While it is popular to approximate the Poisson-Binomial distribution to the Poisson distribution, this approximation is only valid when the input probabilities are small as evident from the bounds on the error defined by Le Cam's theorem~\cite{le1960} below:

\begin{equation}
\sum_{n=0}^\infty \left| P(\Omega_n) - {\frac{\lambda_N^n e^{-\lambda_N}}{n!}} \right| < 2 \sum_{i=1}^N p_i^2. 
\end{equation}

\noindent where $P(\Omega_n)$ gives the probability of $n$ successes in the Poisson-Binomial domain and $\lambda$ is the Poisson parameter.

The Poisson-Binomial distribution has found increasing use in research applications. Shen et al.~\cite{Shen2013} developed a machine learning approach for metabolite identification from large molecular databases such as KEGG and PubChem. The molecular fingerprint vector was treated as Poisson-Binomial distributed and the resulting peak probability was used for candidate retrieval. Similarly, Lai et al.~\cite{Lai2012} developed a statistical model to predict kinase substrates based on phosphorylation site recognition. Importantly, the probability of observing matches to the consensus sequences was calculated using the Poisson-Binomial distribution. Other groups~\cite{Niida2012},~\cite{Cazier2012} have used this distribution to identify genomic aberrations in tumor samples. Since the probability of an aberration event varies across samples, individual DNA base positions are treated as independent Bernoulli trials with unequal success probabilities for ascertaining the likelihood of a genetic aberration at every position in each sample. Following the same reasoning, models to accurately call rare genetic variants~\cite{Zhou2010},~\cite{Wilm2012} have been proposed.

\section{Poisson-Binomial radius semimetric distance}
Before we introduce the PBR distance, we aim to show that feature elements are not identically distributed by necessity.

Let $X$ = $\big($ $X_{1},X_{2},.....,X_{N}$ $\big)$ be an $N$-dimensional random vector where $X_{1},X_{2},.....,X_{N}$ are real-valued random variables. If $X_{i}$ and $X_{j}$ are non-identically distributed, then their distributions $F_{i}$ and $F_{j}$ will differ. Whether this difference exceeds a statistical threshold can be evaluated using the Kolmogorov-Smirnov two sample test (K-S test)~\cite{conover1998}. The K-S test does not make any assumptions about data distributions and can thus be used to probe whether distributions are equal for any arbitrary pair of features. Thus, it has been widely used as a feature selection/weighting procedure to identify significant feature elements in feature vectors~\cite{chen2010},\cite{cieslak2009}. In this study, we used the K-S test to investigate the null hypothesis that a pair of feature elements are drawn from the same distribution. This scenario should be distinguished from the covariate shift, a situation where the feature vectors of training and test data follow different distributions~\cite{bassam2010}. Taking the features for several benchmark data sets which we use in our experiments in the next section, we performed the K-S test with Bonferroni correction on pairwise combinations of feature elements to determine the percentage of feature element pairs found to have significantly different distributions. The experimental results for these data sets with 200,000 pairwise comparisons per data set are reported in Table~\ref{tab:KS_Test}. The majority of K-S tested feature element pairs were found to be significantly different and this result was robust over a range of confidence intervals (95 \%, 99.5 \%, and 99.9 \%).

\renewcommand{\arraystretch}{1.3}
\begin{table}[ht!]
\caption{Results of K-S test (with Bonferroni correction) on feature elements for various data sets with $\alpha$ = 0.05 (95\% confidence), $\alpha$ = 0.005 (99.5\% confidence) and $\alpha$ = 0.001 (99.9\% confidence)}
\centering
\begin{tabular}{cccc}
	\cline{1-4} & \multicolumn{3}{c}{Significantly differently distributed }\\
	& \multicolumn{3}{c}{feature elements (percent)} \\
	\cline{2-4} 
	\multicolumn{1}{c}{Data set} & $\alpha$ = 0.05 & $\alpha$ = 0.005 & $\alpha$ = 0.001\\
	\hline
	FMD~\cite{sharan2014} & 95.57 & 94.85 & 94.43\\
	Kylberg~\cite{kylberg2011} & 99.63 & 99.54 & 99.45\\
	USTB~\cite{yuan2005} & 76.84 & 74.84 & 73.85\\
	IIT Delhi I~\cite{kumar2012} & 86.75 & 85.53 & 84.57\\
	\hline
\end{tabular}
\label{tab:KS_Test}
\end{table}

There is an emerging argument from the machine learning community that feature elements may have dependency for image representation. To take such dependencies into account, we repeated the experiment using the Wilcoxon-signed rank test which makes no assumption of independence. The results of this experiment in Table~\ref{tab:WSR_Test} mirrored that of the K-S test. Median p-values generated by the K-S test for the four data sets were highly significant and ranged from $\leq1.00 \times 10^{-308}$ to $1.79 \times 10^{-26}$. The corresponding range for the Wilcoxon-signed rank test was from $\leq1.00 \times 10^{-308}$ to $9.84 \times 10^{-21}$. These low p-values further suggest that the null hypothesis can be rejected for any reasonable threshold of significance. Taking Tables~\ref{tab:KS_Test} and ~\ref{tab:WSR_Test} in totality, the evidence demonstrates that the identical distribution assumption is not reflective of real-world data, regardless of whether feature elements are assumed to be independent or non-independent.

\renewcommand{\arraystretch}{1.3}
\begin{table}[ht!]
\caption{Results of Wilcoxon-signed rank test (with Bonferroni corrections) on feature elements for various data sets with $\alpha$ = 0.05 (95\% confidence), $\alpha$ = 0.005 (99.5\% confidence) and $\alpha$ = 0.001 (99.9\% confidence)}
\centering
\begin{tabular}{cccc}
	\cline{1-4} & \multicolumn{3}{c}{Significantly differently distributed }\\
	& \multicolumn{3}{c}{feature elements (percent)} \\
	\cline{2-4} 
	\multicolumn{1}{c}{Data set} & $\alpha$ = 0.05 & $\alpha$ = 0.005 & $\alpha$ = 0.001\\
	\hline
	FMD~\cite{sharan2014} & 93.35 & 92.83 & 92.45\\
	Kylberg~\cite{kylberg2011} & 97.35 & 97.13 & 97.01\\
	USTB~\cite{yuan2005} & 79.27 & 77.47 & 76.32\\
	IIT Delhi I~\cite{kumar2012} & 87.72 & 86.72 & 86.03\\
	\hline
\end{tabular}
\label{tab:WSR_Test}
\end{table}

Given the above results, we were naturally motivated to ask if a distribution-agnostic distance measure would actually lead to improved performance for image classification and recognition. From a maximum likelihood perspective, probability distributions play a fundamental role in distance measures by modeling the noise distribution. Besides the well-known Gaussian and exponential distributions which give rise to the $L_2$ (Euclidean) and $L_1$ (Manhattan) distances respectively, practically any distribution (e.g. Cauchy and Gamma-Compound-Laplace) can be used to create an appropriate distance measure~\cite{Sebe2000},\cite{jia2011}. Since all distributions currently used in distance measures do make the identical distribution assumption, we decided to adopt the Poisson-Binomial distribution (PBD) because it is distribution-agnostic. While there are other distributions which also account for non-identical data, such as discrete gamma, discrete burr, discrete Weibull and discrete normal, the Poisson-Binomial distribution is one which is not derived by discretization, a process which leads to information loss~\cite{chakraborty2015}. Further, the discrete nature of PBD is consistent with the nature of pixel intensities, which have also been shown to be discrete random variables governed by quantum mechanics~\cite{hwang2012}. 

Based on the foregoing observations, we propose the PBR distance measure for similarity estimation based on the Poisson-Binomial distribution. In particular, we estimate PBR distance by characterizing the distribution of the difference vector {$E$} (given in the definition below) between two feature vectors ${X}$ and ${Y}$. Although elements in the difference vector {$E$} are not probabilities, we treat them heuristically as such for the purposes of PBR. The Poisson-Binomial distribution can be characterized by its first and second moments, which are respectively its mean and variance. Hence, these summary statistics capture the essence of the difference vector distribution. PBR's premise is to compare the variance of {$E$} to its mean. Two difference vectors ${E_{XY}}$ and ${E_{XZ}}$ arising from three distinct feature vectors ${X}$, ${Y}$ and ${Z}$ will not have the same combination of mean and variance. 

\textbf{Definition.} \textit{Given two N dimensional feature vectors  $ X = (a_1,a_2,a_3,....,a_N) $  and $ Y = (b_1,b_2,b_3,....,b_N) $ with $ E = (e_1,e_2,e_3,....,e_N) $ where $e_i = a_i \ln (\frac{2 a_i}{a_i + b_i}) + b_i \ln (\frac{2 b_i}{a_i + b_i}) $, the Poisson-Binomial Radius distance between X and Y is}

\begin{equation}
{PBR}(X,Y) 	= \displaystyle{\frac{\sigma^2}{N-\mu}} = \frac{{{\displaystyle\sum_{i=1}^{N} e_i (1-e_i)}}}{{N - {\displaystyle\sum_{i=1}^{N} e_i}}}
\end{equation}

PBR is a non-negative function, which satisfies the identity of indiscernibles and symmetry property. However, PBR does not obey the triangle inequality axiom and is thus a semimetric distance.

\begin{figure}[ht!]
	\centering
	\includegraphics[width=12cm]{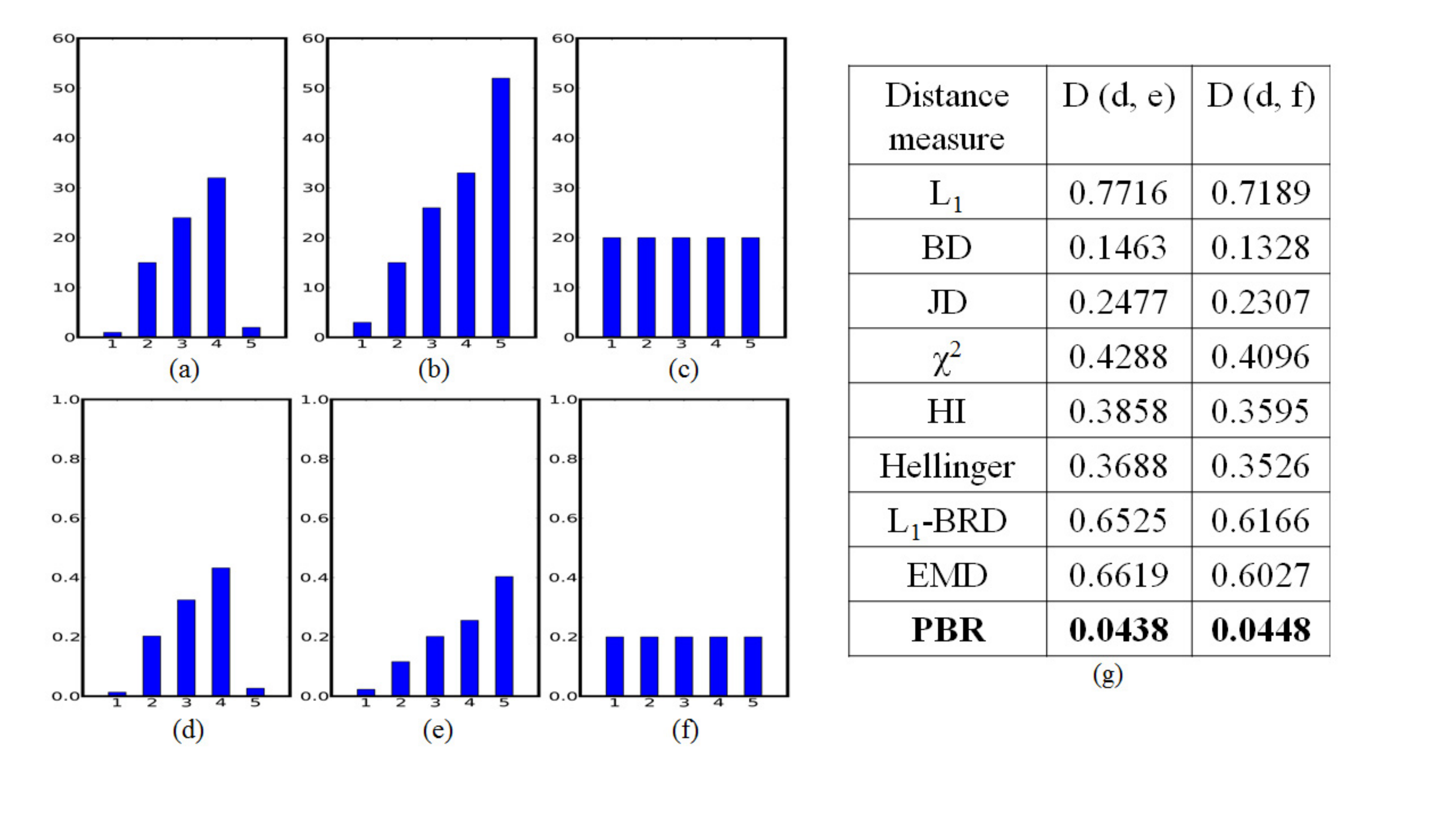}
	\caption{A toy example comparing distances between histograms: (a), (b) and (c) are non-normalized histograms with values [1, 15, 24, 32, 2], [3, 15, 26, 33, 52] and [20, 20, 20, 20, 20], respectively; (d), (e) and (f) are the corresponding normalized histograms; Table (g) shows the distances between the histograms (d) and (e) and between histograms (d) and (f).}
	\label{fig:Toy_data}
\end{figure}

\indent We used toy data to visually compare PBR with commonly used distance measures. In Fig.~\ref{fig:Toy_data}, histograms (a) and (b) represent two images which belong to the same class whereas (c) represents a uniformly distributed reference image. Histograms (d), (e) and (f) are the normalized histograms of (a), (b) and (c) respectively. Table (g) shows that with the exception of PBR, all other distance measures regardless of category (bin-to-bin, cross-bin, intra-cross-bin) erroneously indicate (d) as being more similar to (f) than (e). This is because the Poisson-Binomial variance between histograms (d) and (e) and between histograms (d) and (f) are 0.2079 and 0.2135 respectively, demonstrating that bin-to-bin dispersion was more efficiently accounted for by PBR. As a minor note, PBR produces smaller values relative to the other distance measures because of the large effect of the denominator in Eqn 3.

\section{Kernel-based image classification}
SVM performance depends heavily on the type of kernel employed. The most commonly used kernel is the radial basis function (RBF) because of its universal approximation properties and good generalization capabilities~\cite{wang2004}. Although RBF kernels are already considered generalized, a further level of generalization may be achieved by replacing the $L_2$ norm exponent with a distance measure of choice. Popular choices, which include {$\chi^2$}, HI and {$L_1$-BRD} have achieved high levels of performance when used in the context of RBF kernels~\cite{Sjoberg2013},\cite{hu2014}. Generalized RBFs are considered the gold-standard kernel class in computer vision applications~\cite{zhang2007},\cite{Vempati2010}. In order to enable cross comparisons with current state-of-the-art kernels, we built an RBF-based PBR kernel using the generalized RBF kernel:

\begin{equation}
K_{d-RBF}(X,Y) = e^{-\gamma D(X,Y)}
\end{equation}

where $\gamma$ is a scaling parameter obtained using cross-validation and $D(X,Y)$ is the distance between two feature vectors $X$ and $Y$. Although we do not prove the positive definiteness for the PBR kernel, the PBR kernel has always produced positive definite Gram matrices in all our experiments. Moreover, it is also worth mentioning that non-Mercer kernels such as the EMD kernel work effectively in image classification and recognition tasks~\cite{zhang2007},\cite{chapelle1999}.
\newline
\indent In this work, we compare the performance of PBR distance with BD, JD, {$\chi^2$}, {$L_1$-BRD} and Hellinger distance measures. In addition, we also use linear and polynomial kernels as baseline kernels. Do note that there are state-of-the-art algorithms that perform better than the results reported here. These algorithms are implemented using complex machine learning techniques such as multiple features~\cite{sharan2013}, advanced encoding strategies~\cite{xie2014} and more. The goal of these studies is to improve algorithmic design, particularly in the area of feature representation. In contrast, our main concern is to compare PBR's classification performance with other state-of-the-art distance measures on a level playing field. For this reason, making comparisons to other algorithms is beyond the scope of this paper.

Here, we evaluated the performance of the PBR distance in the following six different classification / recognition applications: texture classification, material recognition, leaf recognition, scene recognition, ear biometrics and category-level image classification.  We used the Brodatz~\cite{brodatz1966}, KTH-TIPS~\cite{hayman2004} and Kylberg~\cite{kylberg2011} data set for texture classification. For material, leaf and scene recognition we used the FMD~\cite{sharan2014}, Swedish Leaf~\cite{soderkvist2001} and MIT Indoor 67~\cite{Quattoni2009} data sets respectively. The Ear biometrics application was tested using USTB~\cite{yuan2005}, IIT Delhi ear databases I and II~\cite{kumar2012}. Category-level image classification was evaluated using the Caltech-101~\cite{fei2007} data set and additionally, a binary image classification task based on the LFW~\cite{Huang2007} and cat data sets~\cite{Zhang2008}. 

\indent For each data set, we randomly split training and testing sets according to protocols recommended by the data set authors. For texture, material and leaf classification, in addition to the standard testing protocol, we evaluated the dependence of performance on the number of training images per class. For ear biometrics, the testing image for each subject was randomly picked and the remaining images were used for training. All experiments were repeated 100 times using a subsampling cross-validation approach except for the MIT Indoor 67 and Caltech-101 data sets. In these two cases, results were based on repeating the subsampling procedure 10 times due to computational complexity. The average accuracy per category was calculated for each individual run. Mean accuracy and standard deviation were reported as the final result. Grayscale intensity values were used for all data sets, even when color images were available. 
\newline
\indent We extracted the Pairwise Rotation Invariant Co-occurrence Local Binary Pattern (PRICoLBP)~\cite{qi2014} feature for all our experiments. PRICoLBP has been shown to be efficient and effective in a variety of applications. The significant attributes of this feature are rotational invariance and effective capture of spatial context co-occurrence information. Particular parameter settings of this feature for individual data sets are described in the next section.
\newline
\indent We used the one-vs-the-rest technique for multi-class SVM classification. SVM hyperparameters such as C, $\gamma$ and degree $d$ were chosen by cross-validation in the training set. For each data set, parameters were obtained for each distance measure separately using the candidate set $log_2$  C $\in$ [-2, 16] and $log_2$  $\gamma$ $\in$ [-4, 8] (with step size 2). These candidate ranges, which can be considered to be extreme RBF parameter values, were employed based on~\cite{albatal2014} where such values were shown to result in superior classification performance in the context of visual descriptor classification. The degree was chosen from the range $d$ $\in$ [1,2,3,4,5] and coefficient was set to 1 for the polynomial kernel.
\newline
\indent In order to identify significant differences between PBR and the other methods, we used the Wilcoxon signed-rank test with Bonferroni correction to control for Type I error~\cite{gama2004}, for each data set. This test is preferred over the resampled paired \textit{t}-test, as the latter tends to underestimate variance, thus inflating Type I error in repeated subsampling cross-validation procedures~\cite{nadeau2003}.
\newline
\indent To further show the generalization capability of the proposed kernel, we estimated the number of support vectors (SVs) for each model chosen by cross-validation for all tested methods in each data set. A classifier which has a large portion of the training data as SVs can be said to have poor generalization performance or to have overfitted the data~\cite{deselaers2010}. 
\newline

\subsection{Texture classification}
\begin{figure*}[ht!]
	\centering
	\begin{subfigure}[b]{0.49\textwidth}
		\includegraphics[width=\textwidth]{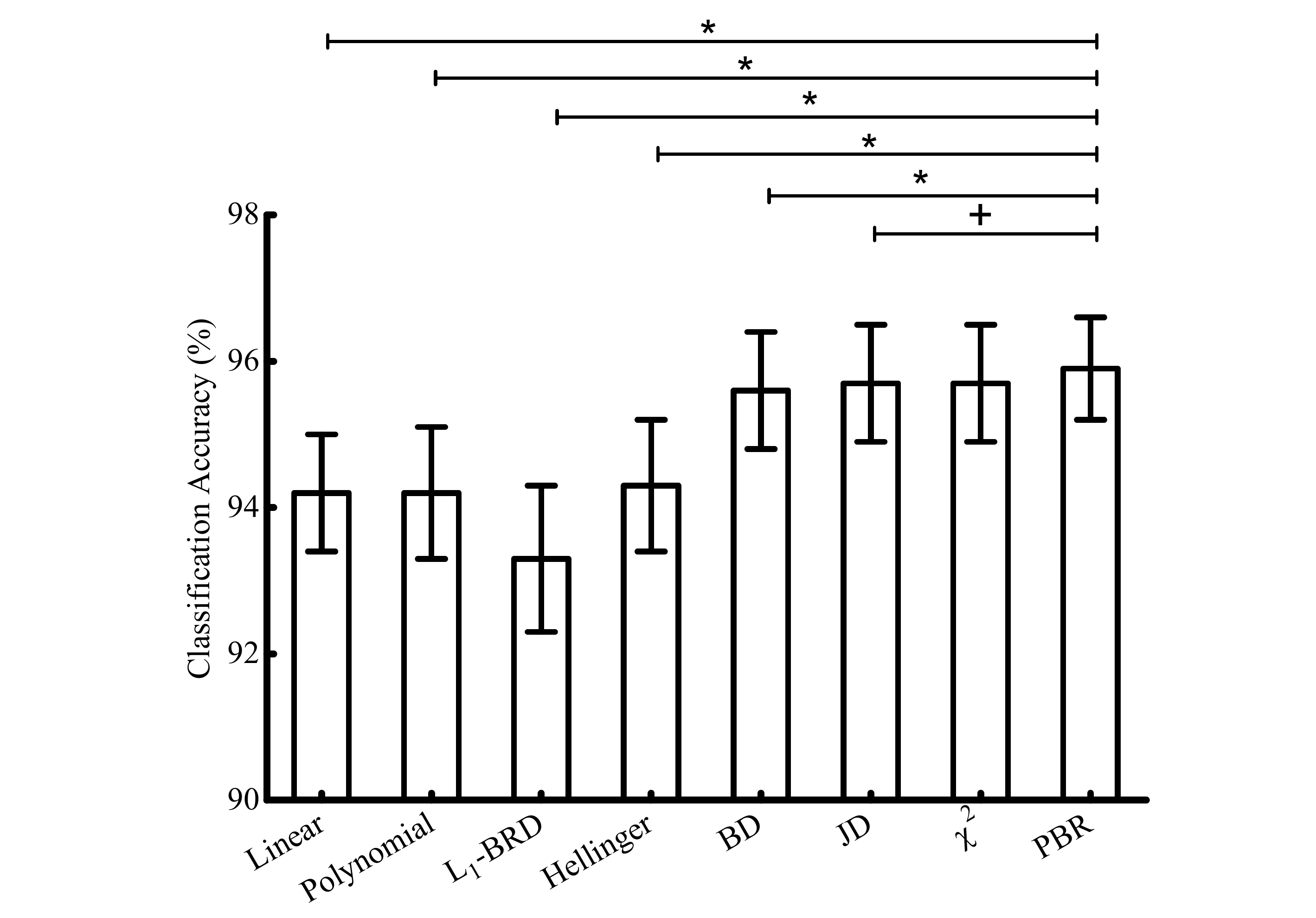}
	\end{subfigure}
	\begin{subfigure}[b]{0.49\textwidth}
		\includegraphics[width=\textwidth]{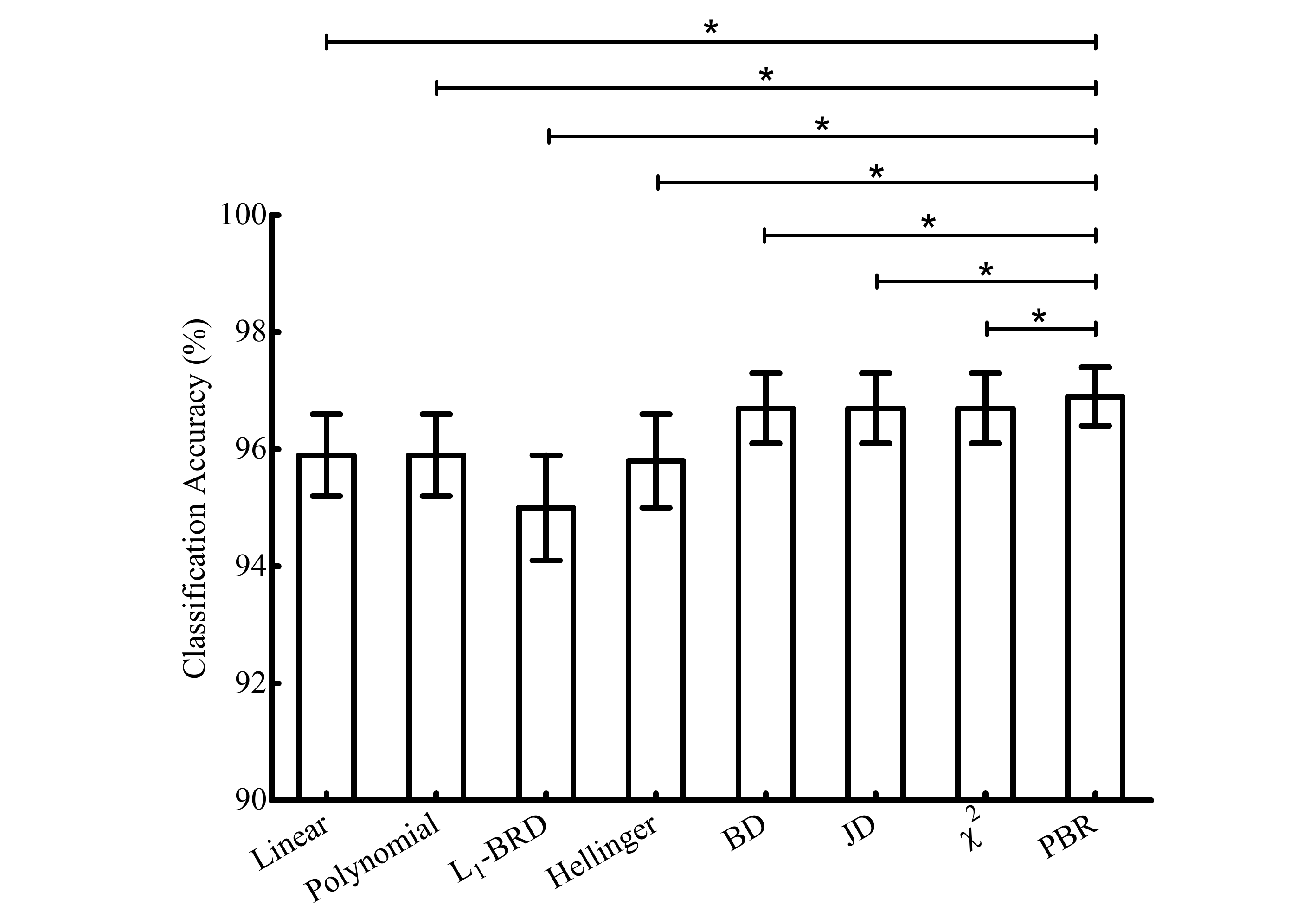}
	\end{subfigure}
\caption{Classification accuracy (percent) for the Brodatz data set using (a) 2 and (b) 3 training images per class. Means and standard deviations are reported. PBR significantly (Bonferroni corrected Wilcoxon-signed rank test) outperforms at $\alpha$ = 0.05 (95\% confidence) indicated by `+' and $\alpha$ = 0.005 (99.5\% confidence) indicated by `$\ast$'.}
\label{fig:Brodatz}
\end{figure*}

\begin{figure*}[ht!]
	\centering
	\begin{subfigure}[b]{0.49\textwidth}
		\includegraphics[width=\textwidth]{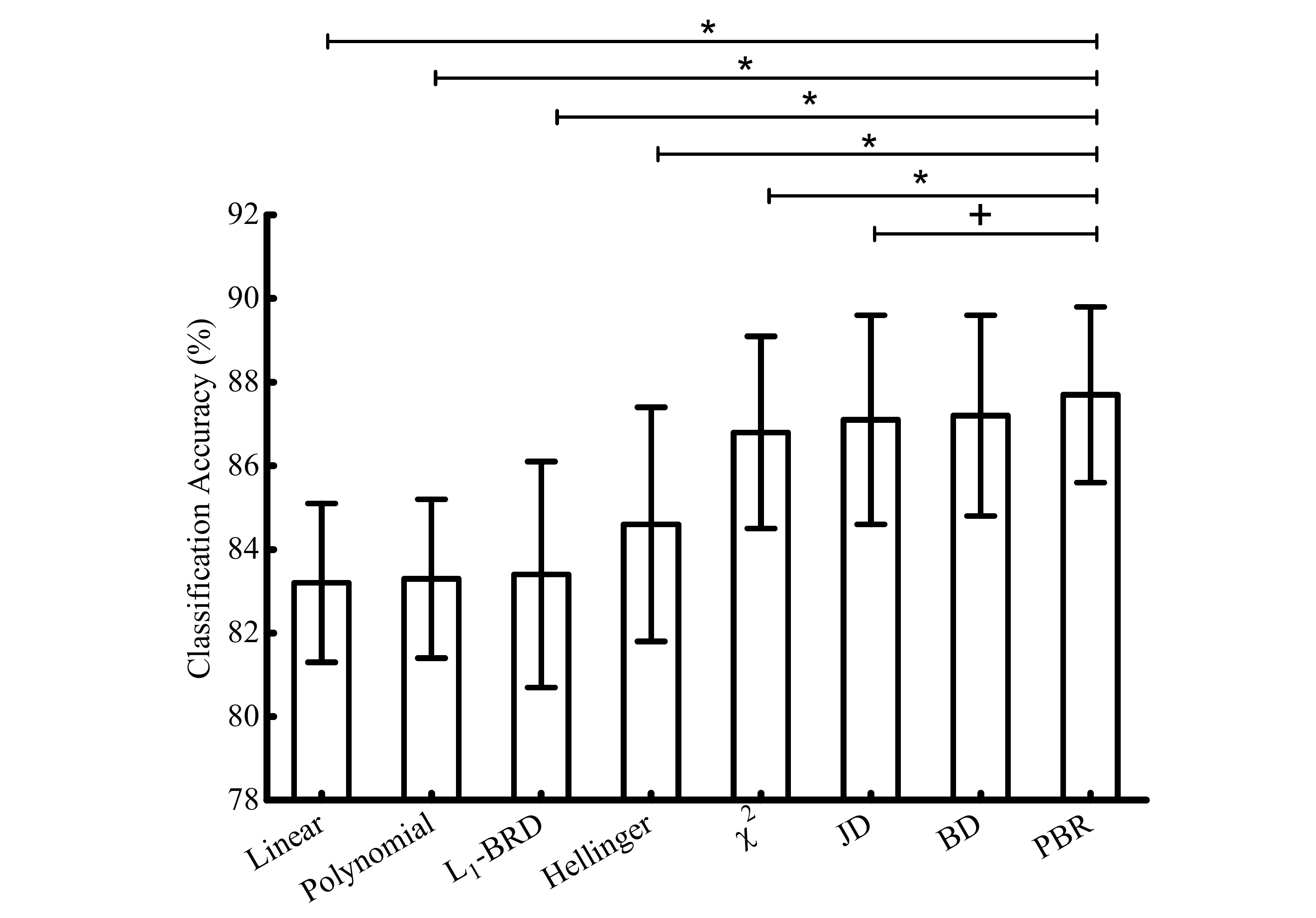}
	\end{subfigure}
	\begin{subfigure}[b]{0.49\textwidth}
		\includegraphics[width=\textwidth]{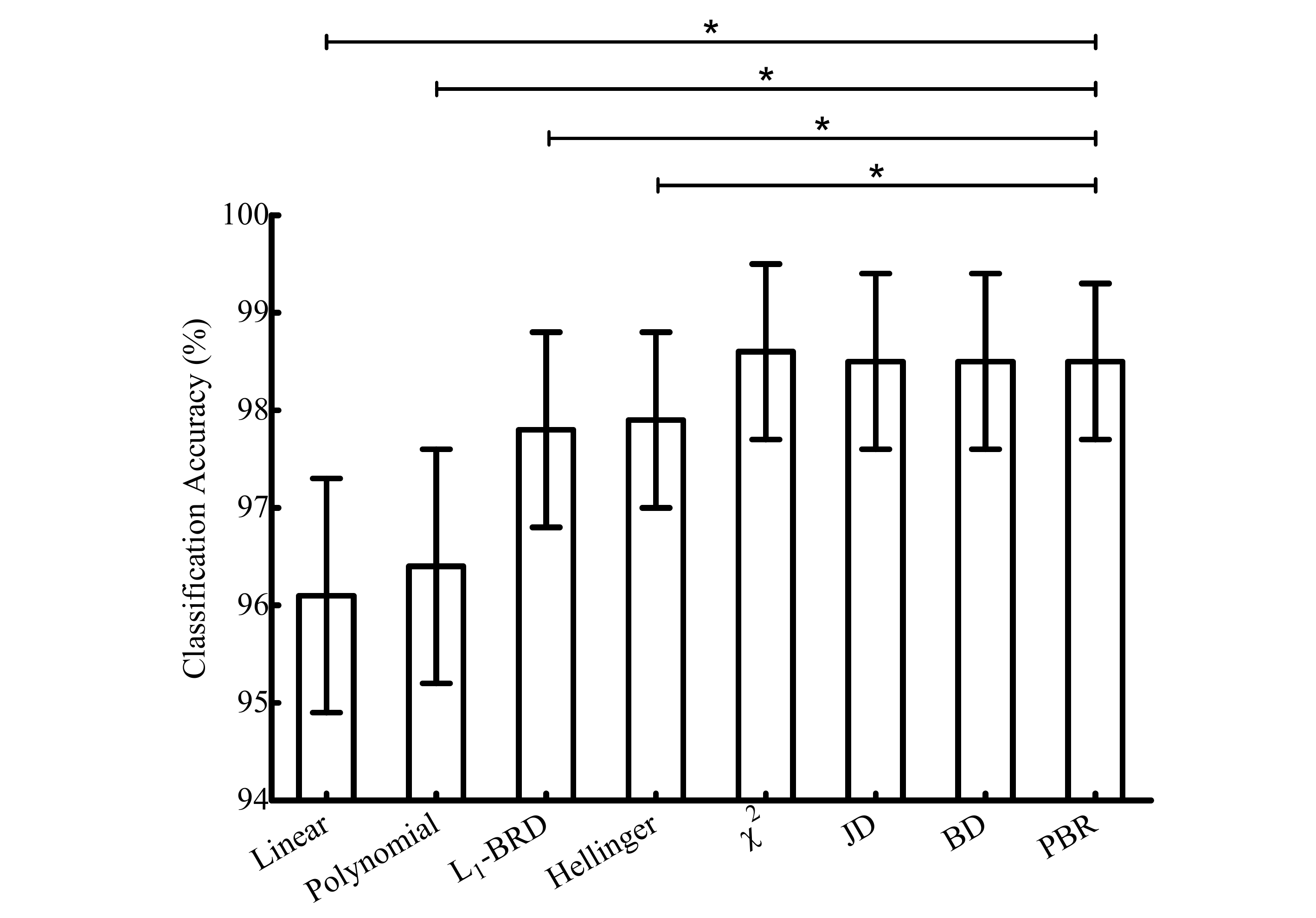}
	\end{subfigure}
\caption{Classification accuracy (percent) for the KTH-TIPS data set using (a) 10 and (b) 40 training images per class. Means and standard deviations are reported. PBR significantly (Bonferroni corrected Wilcoxon-signed rank test) outperforms at $\alpha$ = 0.05 (95\% confidence) indicated by `+' and $\alpha$ = 0.005 (99.5\% confidence) indicated by `$\ast$'.}
\label{fig:KTHTIPS}
\end{figure*}

\begin{figure*}[ht!]
	\centering
	\begin{subfigure}[b]{0.49\textwidth}
		\includegraphics[width=\textwidth]{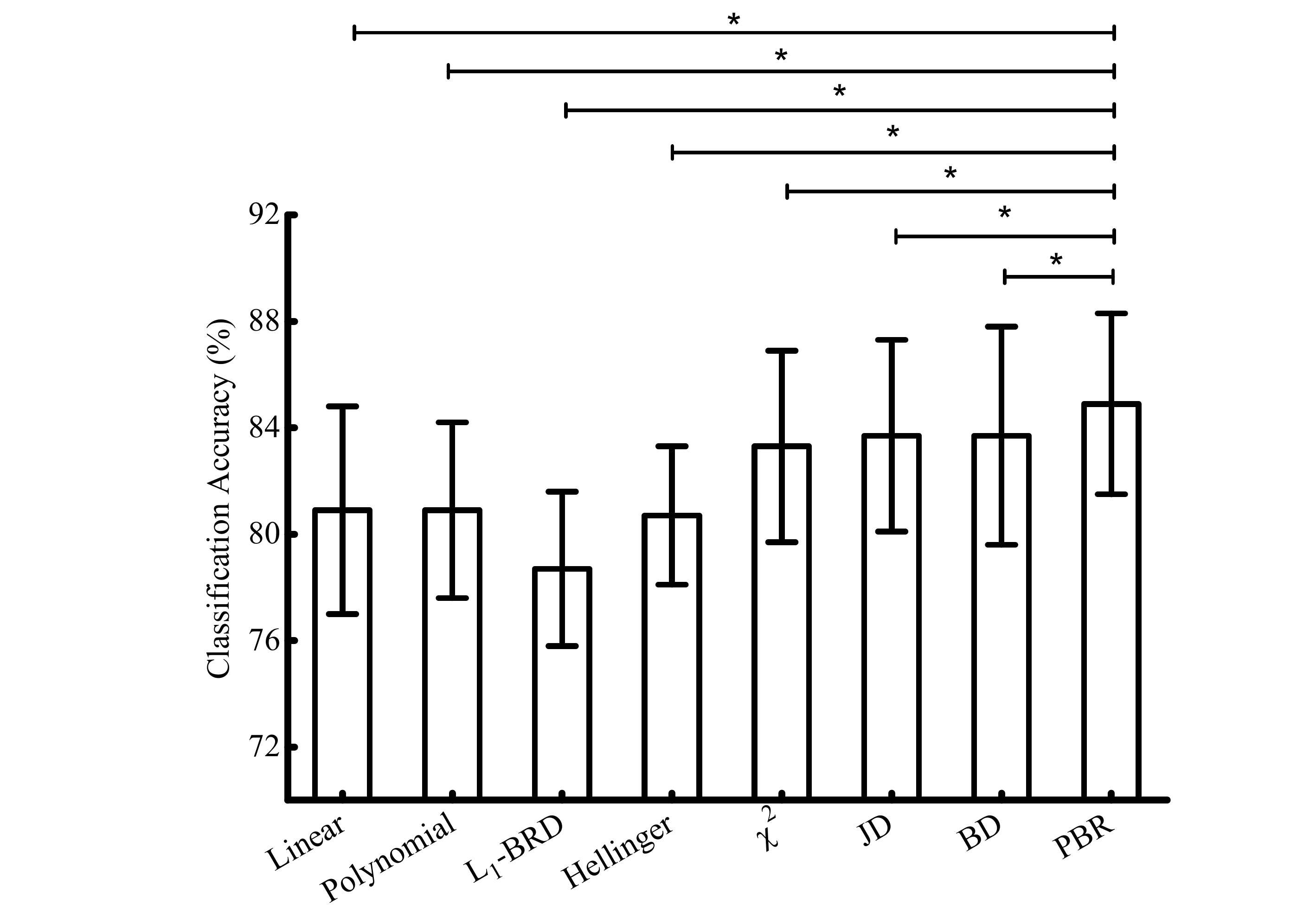}
	\end{subfigure}
	\begin{subfigure}[b]{0.49\textwidth}
		\includegraphics[width=\textwidth]{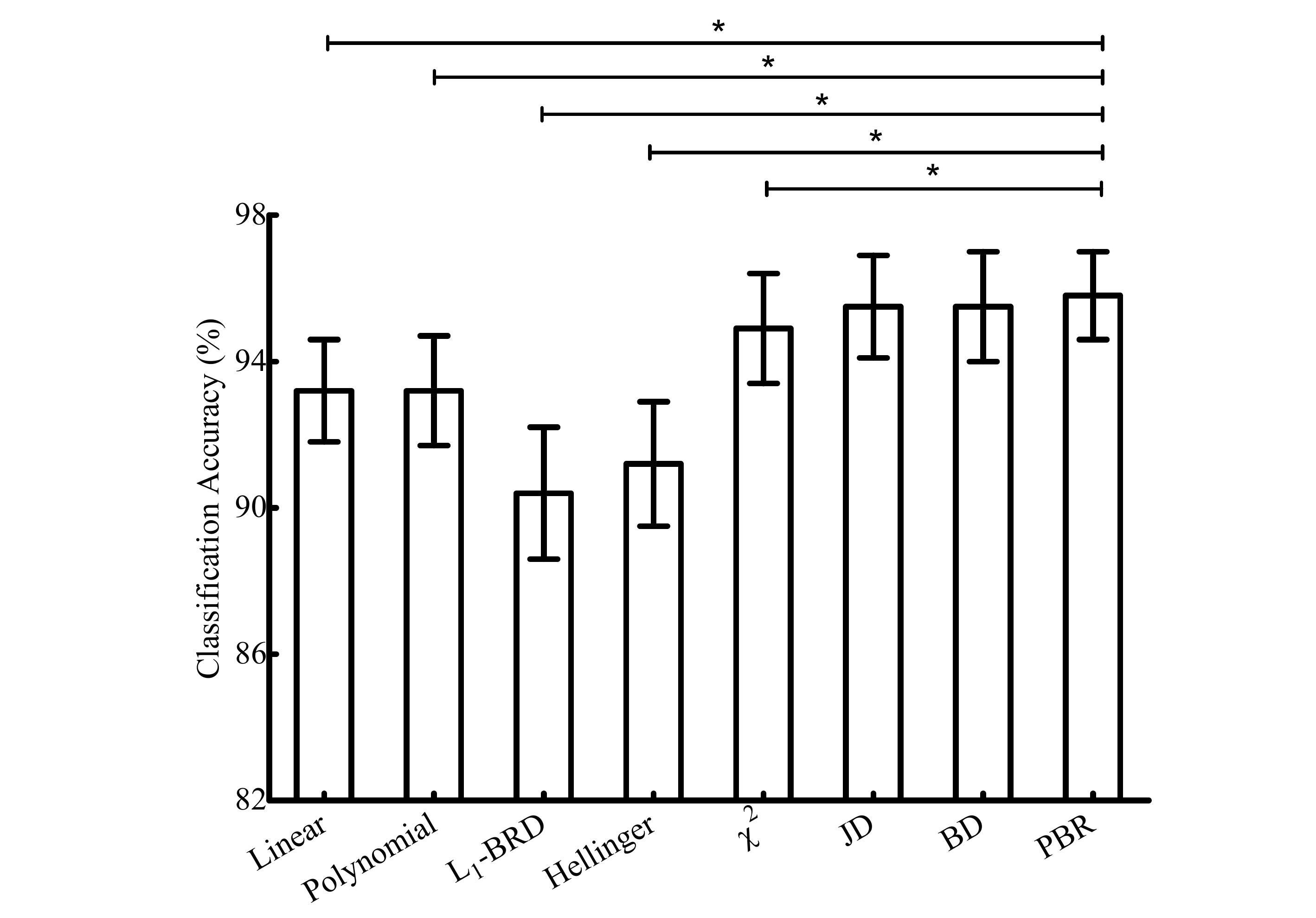}
	\end{subfigure}
\caption{Classification accuracy (percent) for the Kylberg data set using (a) 2 and (b) 5 training images per class. Means and standard deviations are reported. PBR significantly (Bonferroni corrected Wilcoxon-signed rank test) outperforms at $\alpha$ = 0.05 (95\% confidence) indicated by `+' and $\alpha$ = 0.005 (99.5\% confidence) indicated by `$\ast$'.}
\label{fig:Kylberg}
\end{figure*}

The Brodatz album is a popular benchmark texture data set which contains 111 different texture classes. Each class comprises of one image divided into nine non-overlapping sub-images.
\newline
\indent The KTH-TIPS data set consists of 10 texture classes, with 81 images per class. These images demonstrate high intra-class variability because they are captured at nine scales, under three different illumination directions and with three different poses.
\newline
\indent The Kylberg data set has 28 texture classes of 160 unique samples per class. The classes are homogeneous in terms of scale, illumination and directionality. We used the {\textit{without}} rotated texture patches version of the data set.
\newline
\indent The $2_{a}$ template configuration of PRICoLBP was used to produce a 1,180 dimensional feature for all the above data sets. Figs.~\ref{fig:Brodatz},~\ref{fig:KTHTIPS} and~\ref{fig:Kylberg} show the classification accuracy for the Brodatz, KTH-TIPS and Kylberg data sets.

From the results, we observe that:

$\bullet$ PBR outperforms existing semi-metric distances (BD, JD and {$\chi^2$}) which in turn perform better than distance metrics ($L_1$-BRD and Hellinger) and baseline kernels.
\newline
\indent $\bullet$ PBR generally outperforms other methods in a significant manner when the training set is small. The only exceptions are {$\chi^2$} in Brodatz and BD in KTH-TIPS. As the number of training images increases, other methods converge in performance to PBR.

\subsection{Material recognition}

\begin{figure}[!ht]
	\centering
	\includegraphics[width=6cm]{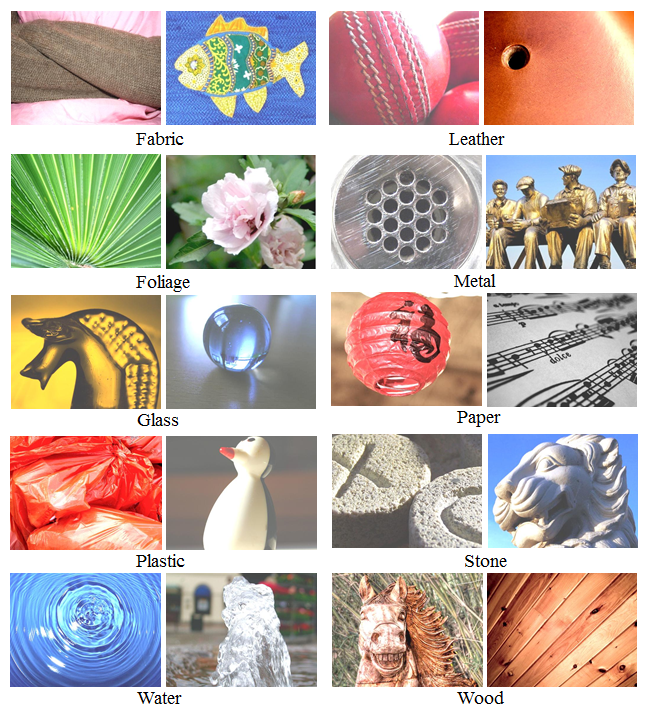}
	\caption{Sample images of FMD data set}
	\label{fig:FMD_Samples}
\end{figure}

The Flickr Material Database (FMD) is a recently published benchmark data set for material recognition. The images in this database are manually selected from Flickr photos and each image belongs to one of 10 common material categories, including fabric, foliage, glass, leather, metal, paper, plastic, stone, water, and wood. Samples images are shown in Fig.~\ref{fig:FMD_Samples}. Each category includes 100 images (50 close-up views and 50 object-level views) which capture the large diversity in appearance of real-world materials. In particular, these images are defined by large intra-class variations in scale, pose and illumination. All images are associated with segmentation masks which describe the location of the object. We use these masks to extract PRICoLBP only from the object regions. Feature extraction was performed using PRICoLBP's 6 template configuration, yielding a 3,540 dimensional feature.

\begin{figure*}[!ht]
	\centering
	\begin{subfigure}[b]{0.49\textwidth}
		\includegraphics[width=\textwidth]{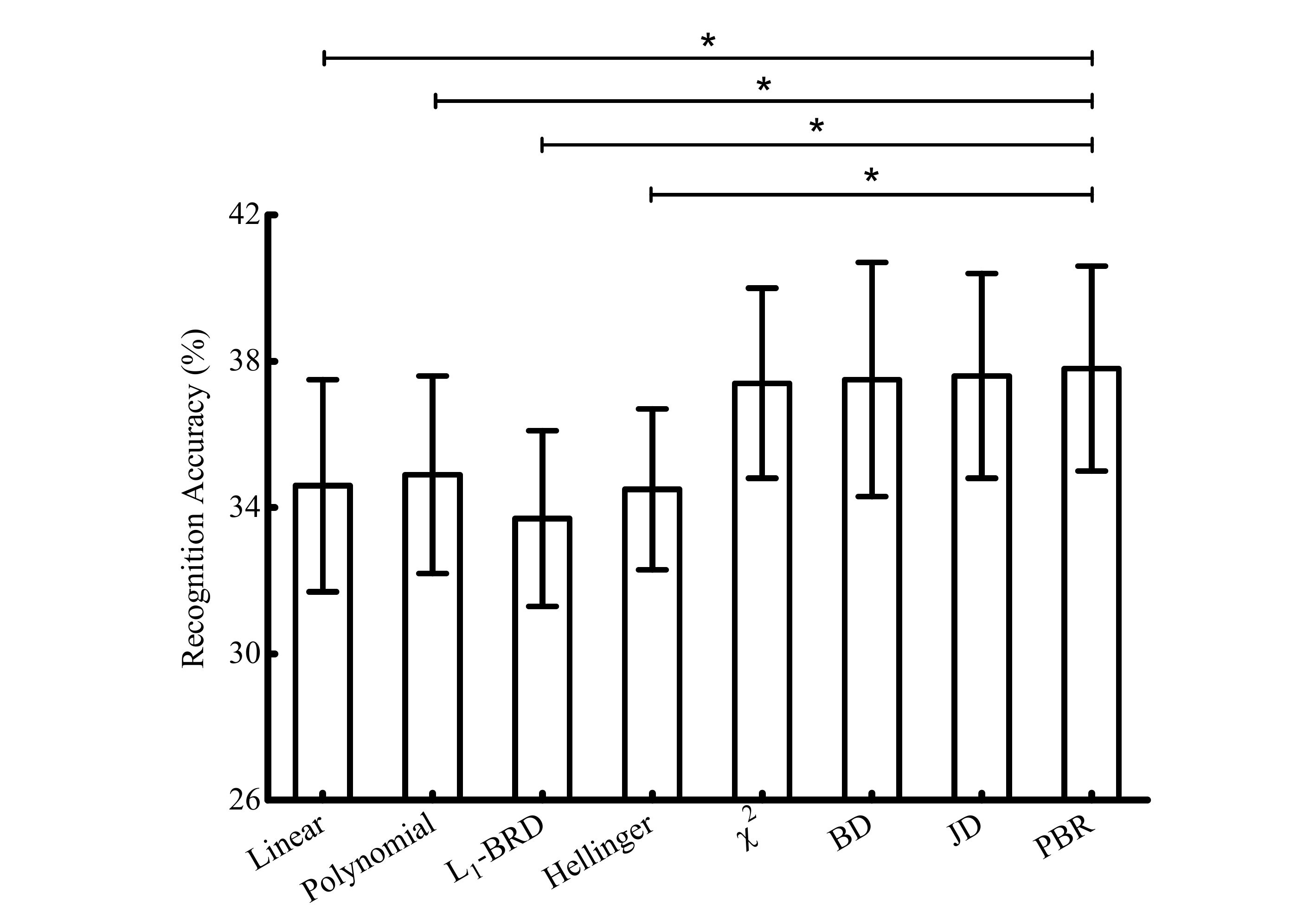}
	\end{subfigure}
	\begin{subfigure}[b]{0.49\textwidth}
		\includegraphics[width=\textwidth]{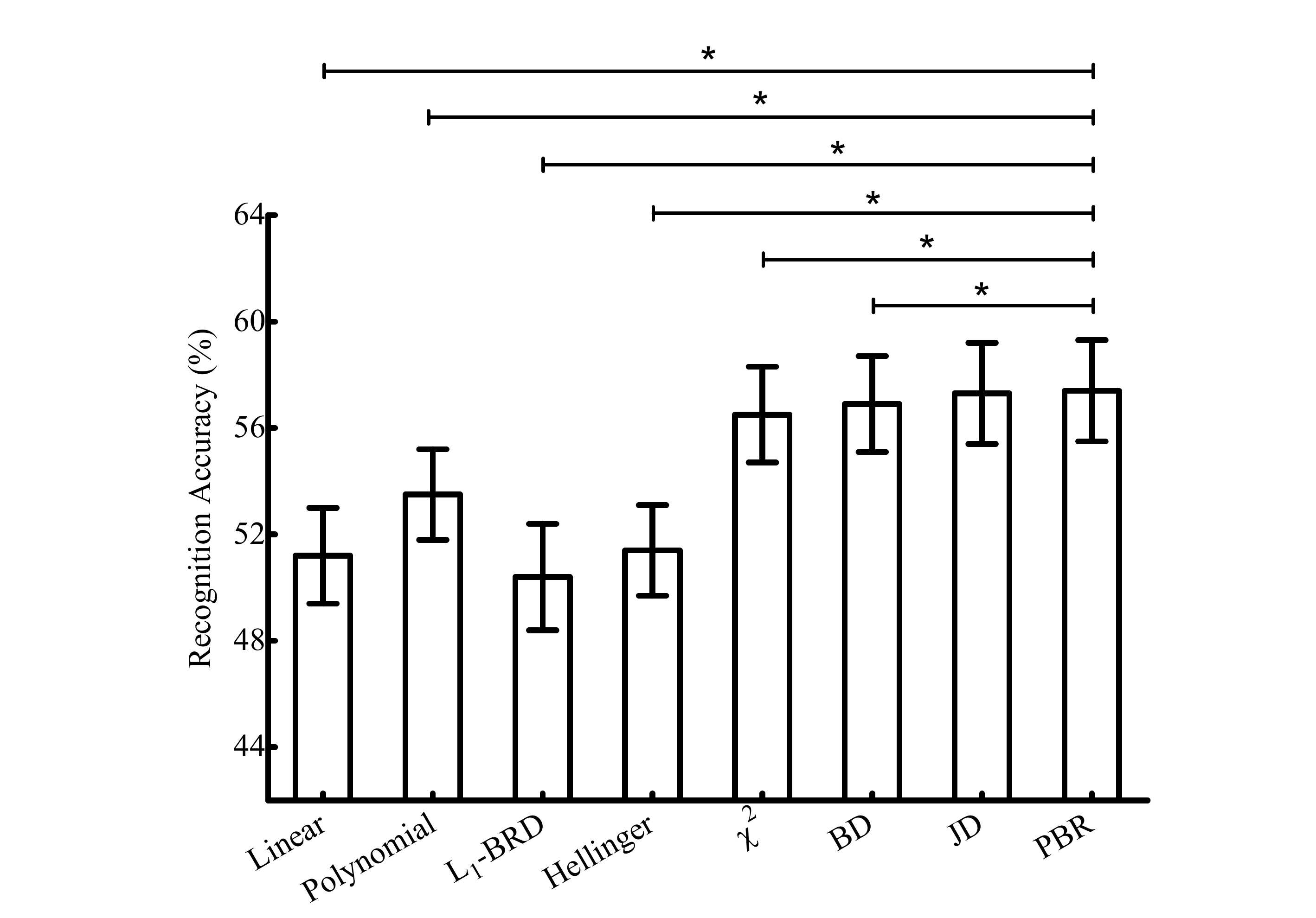}
	\end{subfigure}
\caption{Recognition accuracy (percent) for the FMD data set using (a) 10 and (b) 50 training images per class. Means and standard deviations are reported. PBR significantly (Bonferroni corrected Wilcoxon-signed rank test) outperforms at $\alpha$ = 0.05 (95\% confidence) indicated by `+' and $\alpha$ = 0.005 (99.5\% confidence) indicated by `$\ast$'.}
\label{fig:FMD}
\end{figure*}

Fig.~\ref{fig:FMD} shows recognition rates for the FMD data set. For the case of 10 training images per class, we observe that PBR significantly outperforms (99.5\% confidence) baseline kernels and distance metrics ($L_1$-BRD and Hellinger) while achieving comparable results for the others. When the number of training images per class is raised to 50, PBR significantly outperforms (99.5\% confidence) all other methods with the exception of JD.
\subsection{Leaf recognition}

\begin{figure}[ht!]
	\centering
	\includegraphics[width=12cm]{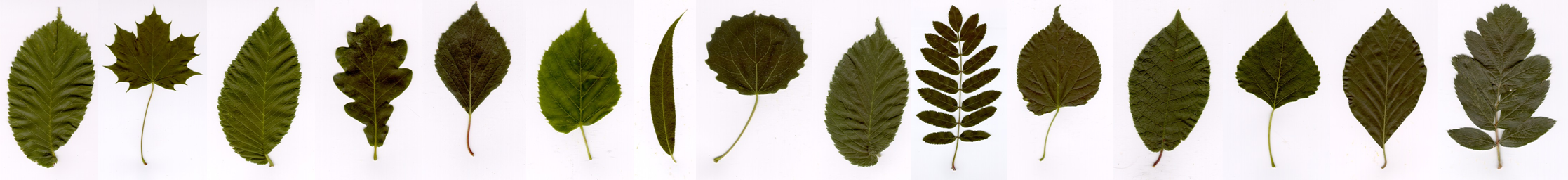}
	\caption{Example images from Swedish leaf dataset, one image per species}
	\label{fig:leaves_samples}
\end{figure}

The Swedish leaf data set contains 15 different Swedish tree species, each represented by 75 images. The hallmark of this data set is that images exhibit high inter-class similarity in combination with intra-class geometric and photometric variation (Fig.~\ref{fig:leaves_samples}).

\begin{figure*}[ht!]
	\centering
	\begin{subfigure}[b]{0.49\textwidth}
		\includegraphics[width=\textwidth]{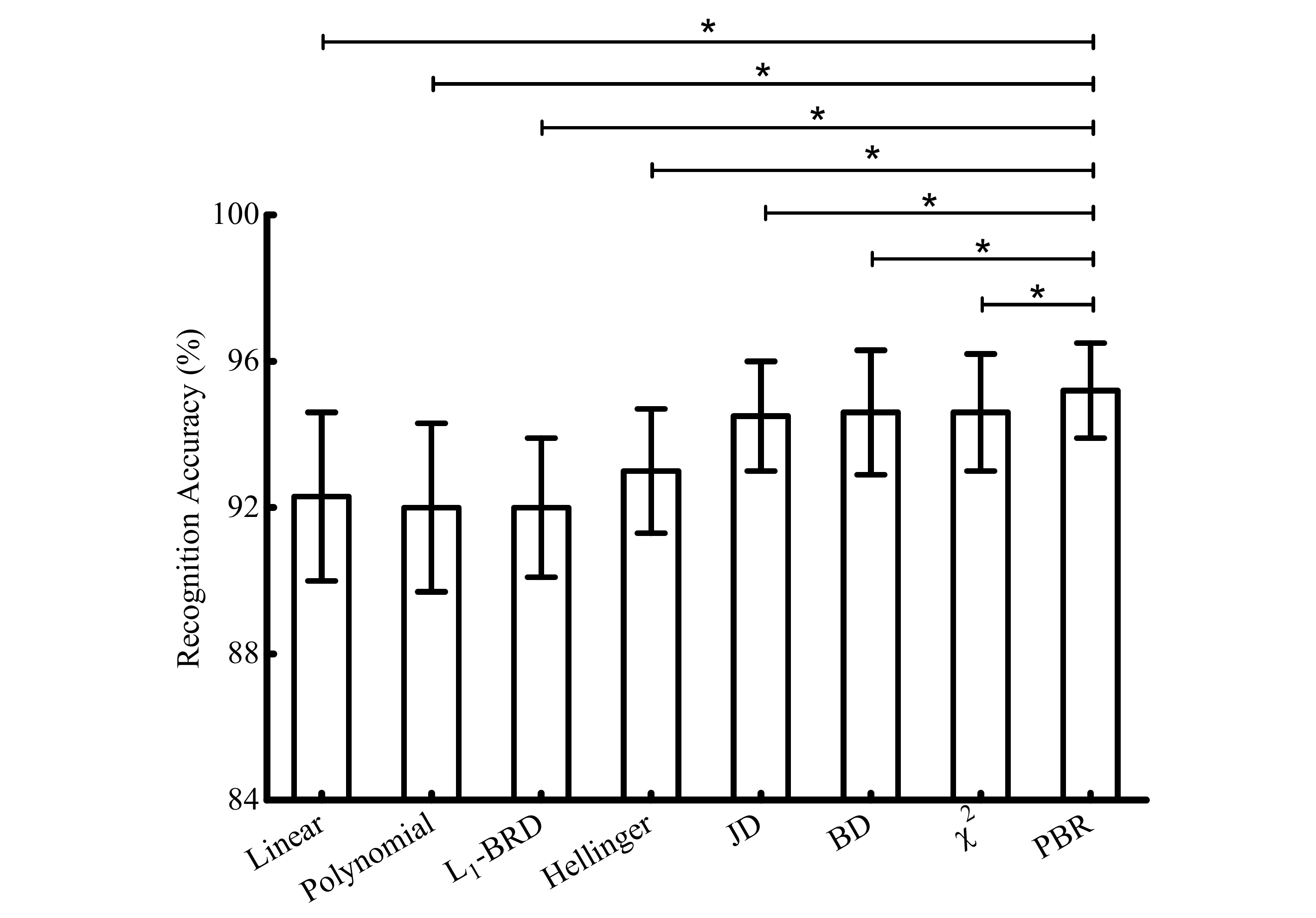}
	\end{subfigure}
	\begin{subfigure}[b]{0.49\textwidth}
		\includegraphics[width=\textwidth]{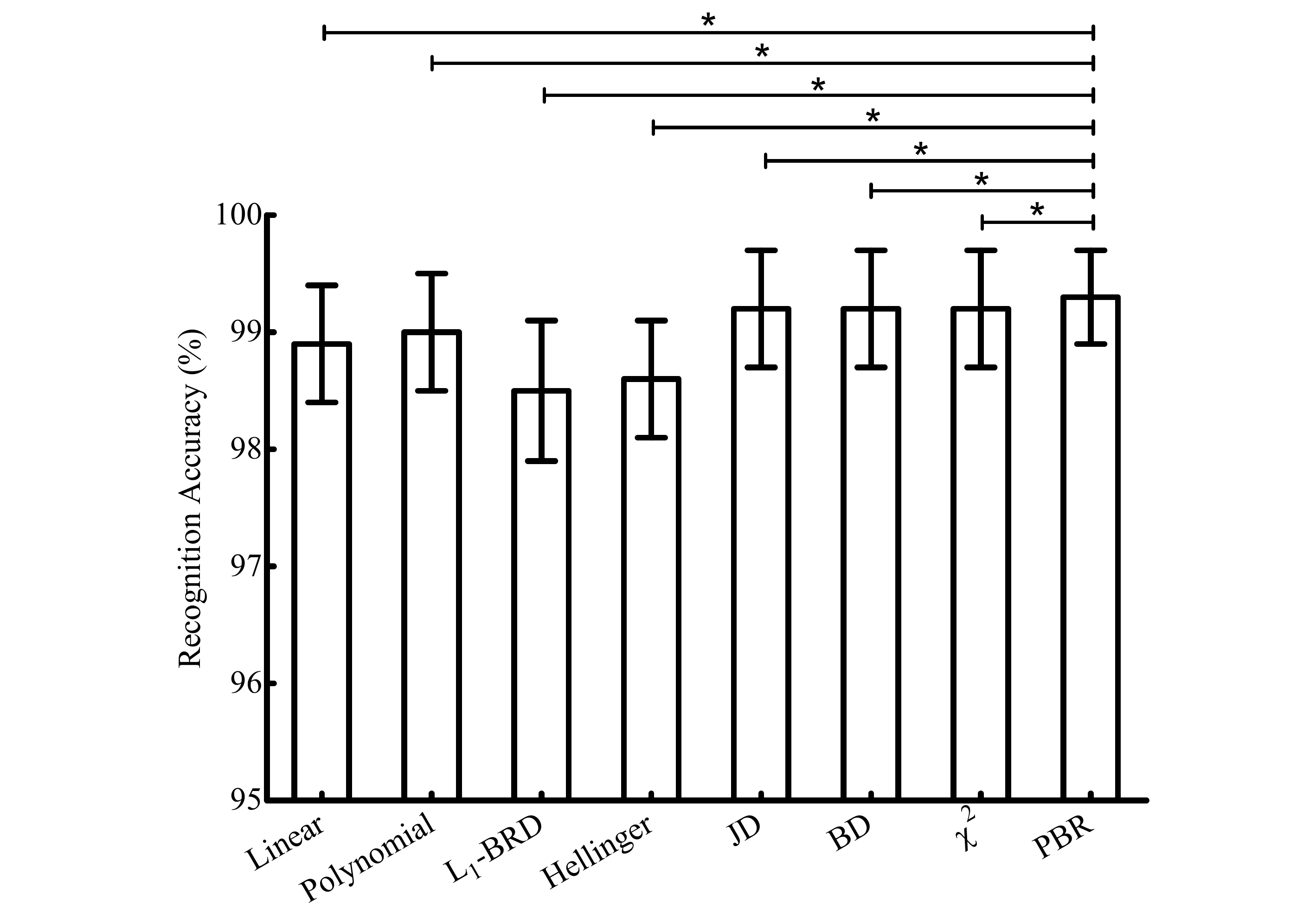}
	\end{subfigure}
\caption{Recognition accuracy (percent) for the Swedish leaf data set using (a) 5 and (b) 25 training images per class. Means and standard deviations are reported. PBR significantly (Bonferroni corrected Wilcoxon-signed rank test) outperforms at $\alpha$ = 0.05 (95\% confidence) indicated by `+' and $\alpha$ = 0.005 (99.5\% confidence) indicated by `$\ast$'.}
\label{fig:Swedish_Leaf}
\end{figure*}

\indent We used the same setting of PRICoLBP as for the texture data set. We did not use the spatial layout prior information of the leaves. Experimental results for 5 and 25 training images per class are shown in Fig.~\ref{fig:Swedish_Leaf}. We observe that PBR significantly outperforms (99.5\% confidence) all other methods for both 5 and 25 training images per class.

\subsection{Scene recognition}
\begin{figure}[!ht]
	\centering
	\includegraphics[width=9cm]{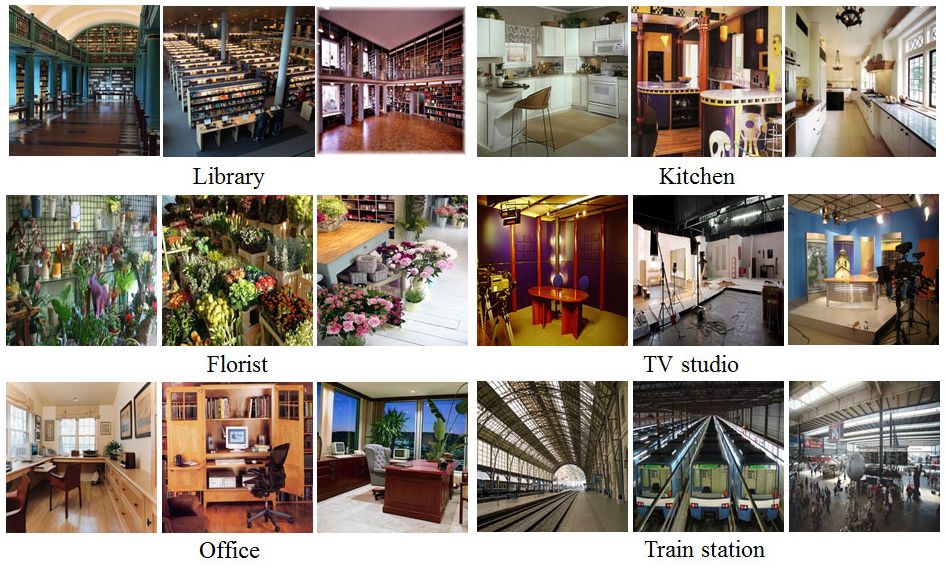}
	\caption{Sample images from six categories of MIT Indoor 67 data set}
	\label{fig:MIT_Samples}
\end{figure}

The MIT Indoor 67 data set contains a total of 15,620 indoor scene images in 67 different categories. The images in this data set were collected from Google, Altavista, Flickr and the LabelMe data set. Sample images are shown in Fig.~\ref{fig:MIT_Samples}. The number of images per category ranges from 101 to 734. These images are of different resolutions, hence we resized the images to have a maximum dimension of 400 pixels while maintaining the aspect ratio. We used 80 images per class for training and 20 images per class for testing.

\begin{figure*}[!hb]
\centering
\includegraphics[width=\textwidth]{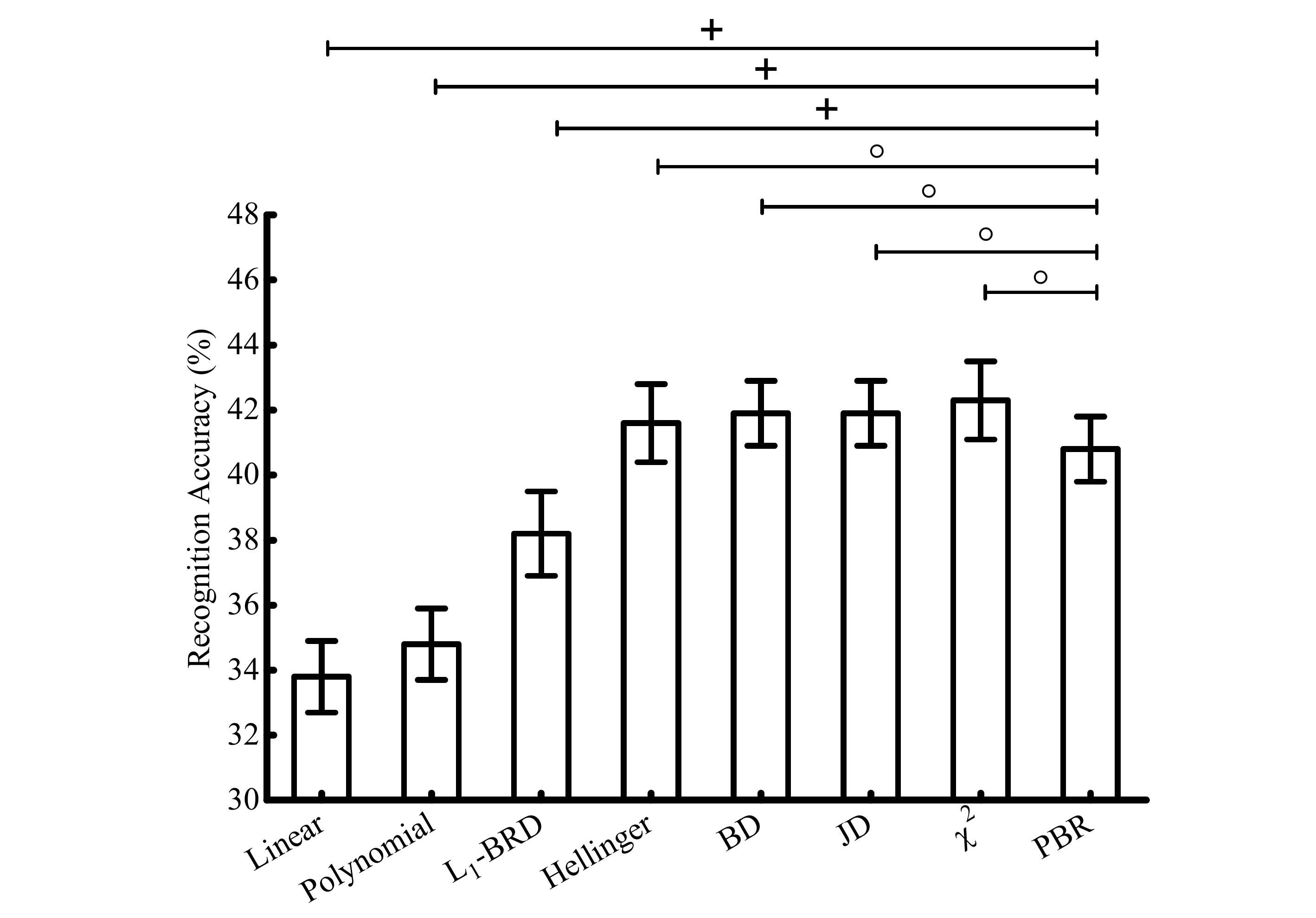}
\caption{Recognition accuracy (percent) for the MIT Indoor 67 data set using 80 training images per class. Means and standard deviations are reported. A `+' or '$\circ$' means that either PBR or the corresponding method significantly (Bonferroni corrected Wilcoxon-signed rank test) outperforms with $\alpha$ = 0.05 (95\% confidence).}
\label{fig:MIT_Indoor}
\end{figure*}

\indent We used the 6 template configuration of PRICoLBP which produces a 3,540 dimensional feature. Recognition performance results in Fig~\ref{fig:MIT_Indoor} show that PBR outperforms baseline kernels and {$L_1$-BRD} with statistically significance (95\% confidence) and achieves comparable performance with respect to other methods.

\subsection{Ear biometrics}
\begin{figure}[!ht]
	\centering
	\includegraphics[width=9cm]{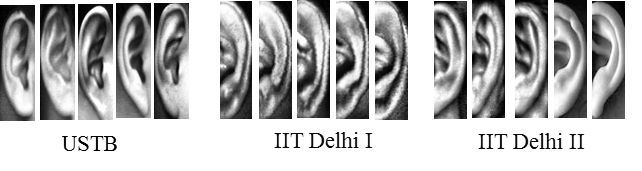}
	\caption{Enhanced ear image samples from five subjects of USTB, IIT Delhi I and II data sets}
	\label{fig:ear_samples}
\end{figure}

\begin{figure*}[!hb]
\centering
\includegraphics[width=\textwidth]{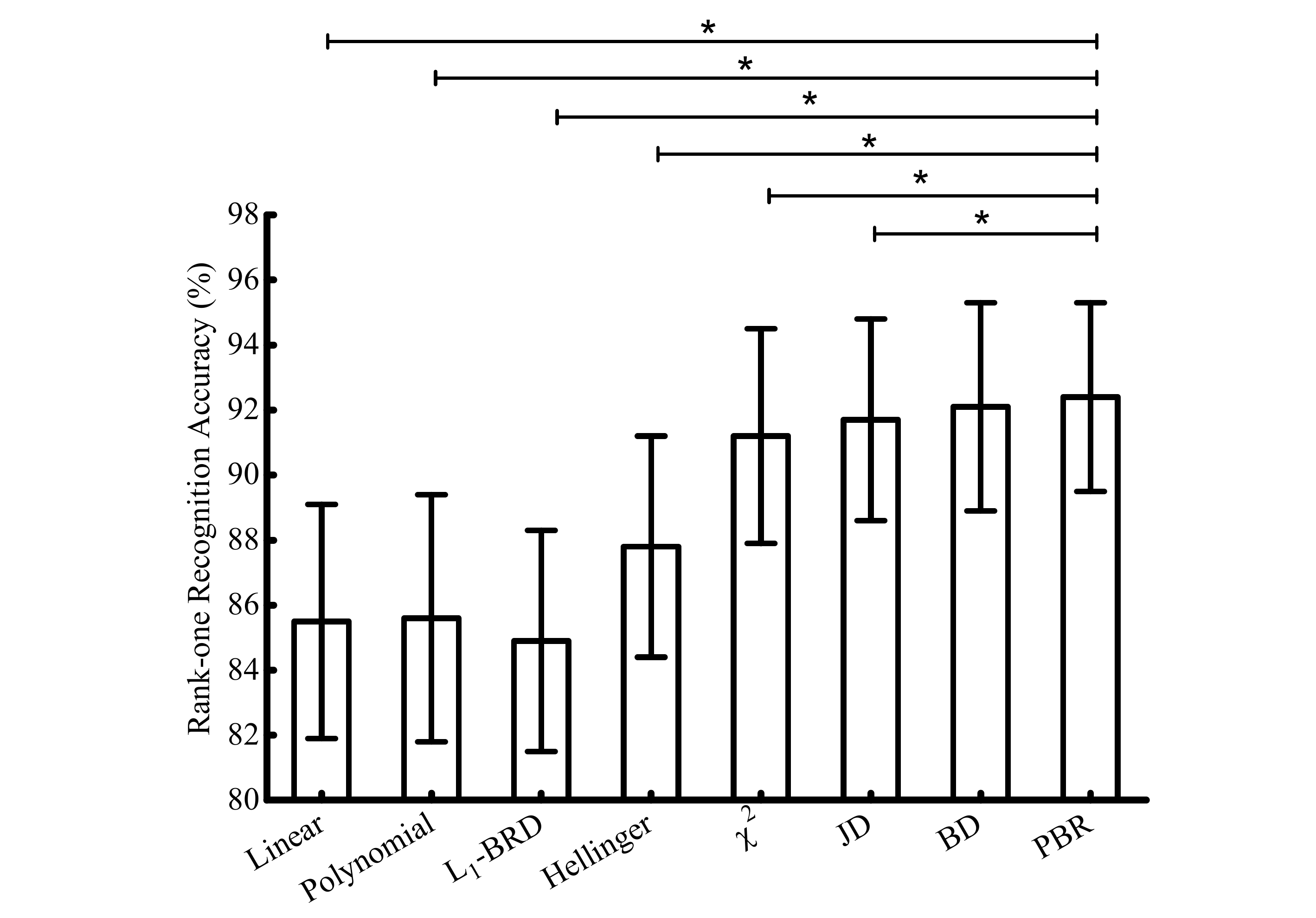}
\caption{Rank-one recognition accuracy (percent) for the USTB data set. Means and standard deviations are reported. A `$\ast$' means that PBR significantly (Bonferroni corrected Wilcoxon-signed rank test) outperforms with $\alpha$ = 0.005 (99.5\% confidence).}
\label{fig:USTB}
\end{figure*}

\begin{figure*}[ht!]
	\centering
	\begin{subfigure}[b]{0.49\textwidth}
		\includegraphics[width=\textwidth]{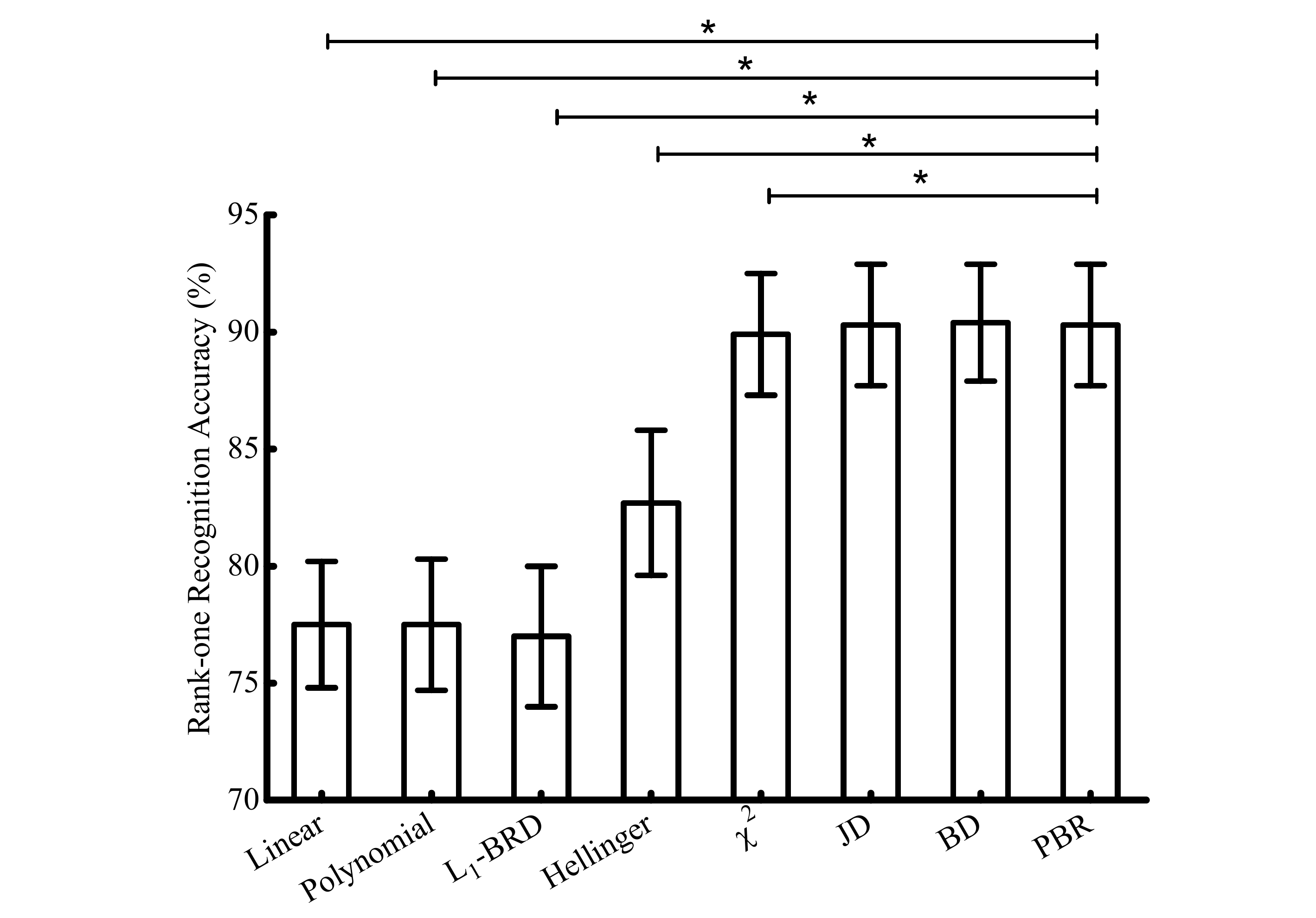}
	\end{subfigure}
	\begin{subfigure}[b]{0.49\textwidth}
		\includegraphics[width=\textwidth]{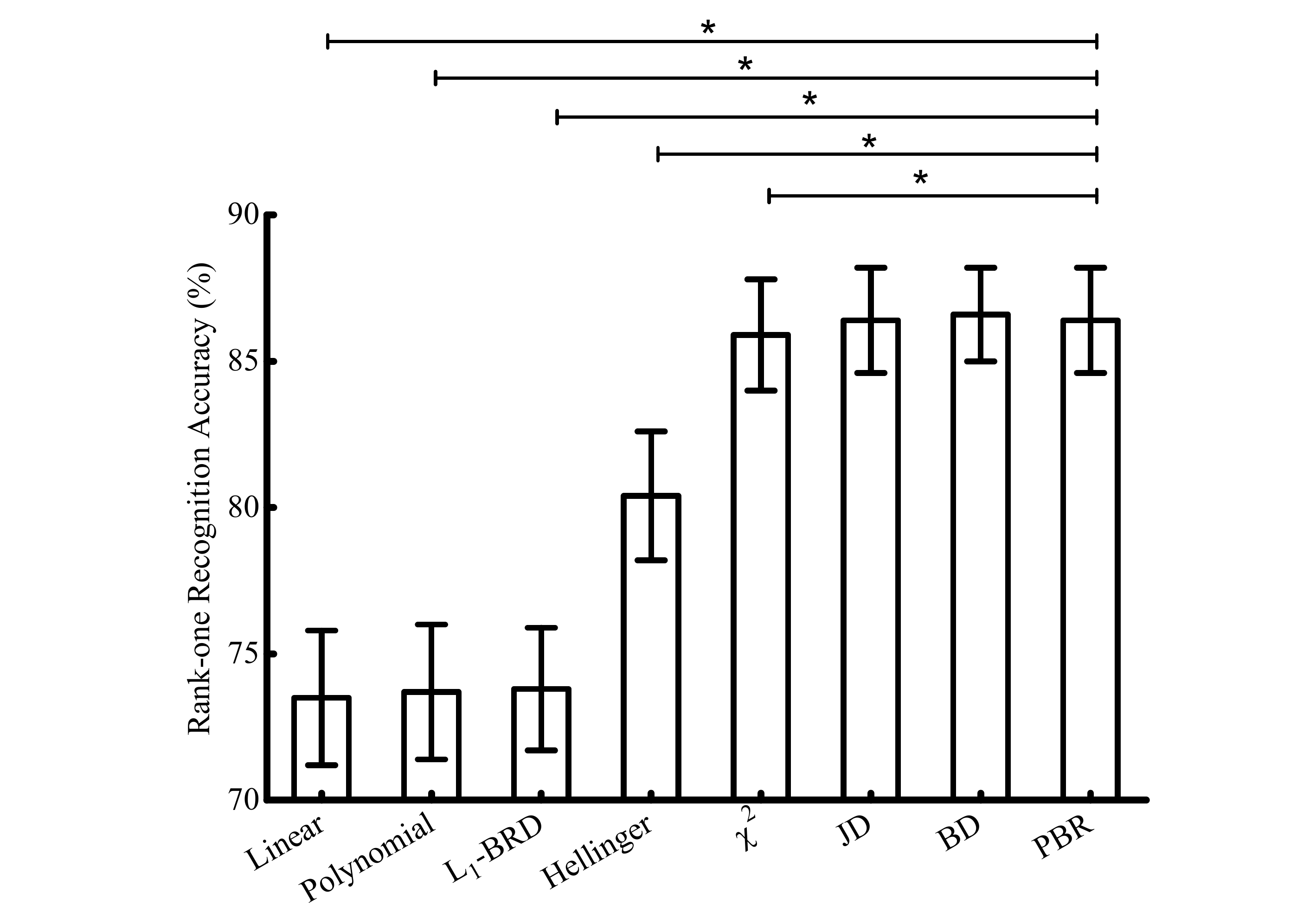}
	\end{subfigure}
\caption{Rank-one recognition accuracy (percent) for the IIT Delhi (a) I and (b) II data sets. Means and standard deviations are reported. A `$\ast$' means that PBR significantly (Bonferroni corrected Wilcoxon-signed rank test) outperforms with $\alpha$ = 0.005 (99.5\% confidence).}
\label{fig:IIT}
\end{figure*}

The IIT Delhi ear data sets I and II consist of 493 ear images from 125 subjects and 793 ear images from 221 subjects respectively. The number of ear images for each subject in these data sets vary from 3 to 6. 
\newline
\indent We also used a similar ear data set from USTB containing 60 subjects and a total of 185 ear images. In this data set, 55 subjects are represented with 3 images and the remaining with 4 images. 
\newline
\indent For the ear recognition application, we used histogram equalization as a pre-processing step. Sample images from the above mentioned data sets after histogram equalization are shown in Fig.~\ref{fig:ear_samples}. We then extracted the PRICoLBP feature using the 6 template configuration. We evaluated performance using rank-one recognition accuracy. The mean and standard deviation of the rank-one recognition rate for different methods are shown in Figs.~\ref{fig:USTB} and ~\ref{fig:IIT}. 
\newline
\indent From Fig.~\ref{fig:USTB}, we can see that PBR works best on the USTB data set. We further observe from Fig.~\ref{fig:IIT} that PBR places second on the IIT Delhi data sets and significantly outperforms (99.5\% confidence) all other methods except BD and JD.

\begin{figure}[ht]
	\centering
	\includegraphics[width=9cm]{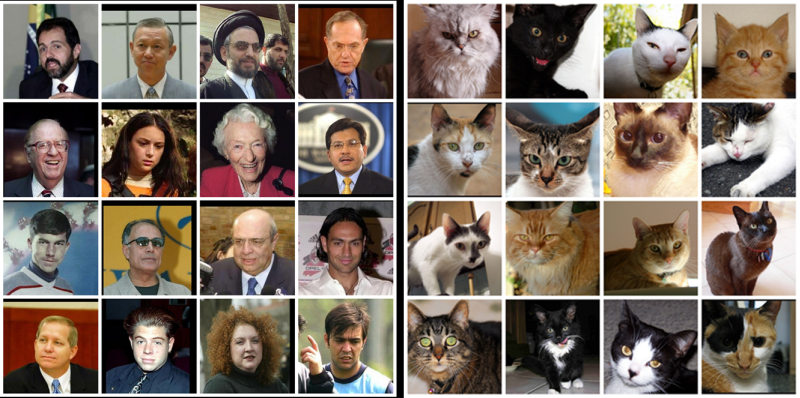}
	\caption{Sample images from LFW (left) and cat (right) data sets}
	\label{fig:Human_cats}
\end{figure}

\subsection{Category-level image classification}
\begin{figure*}[ht!]
	\centering
	\begin{subfigure}[b]{0.49\textwidth}
		\includegraphics[width=\textwidth]{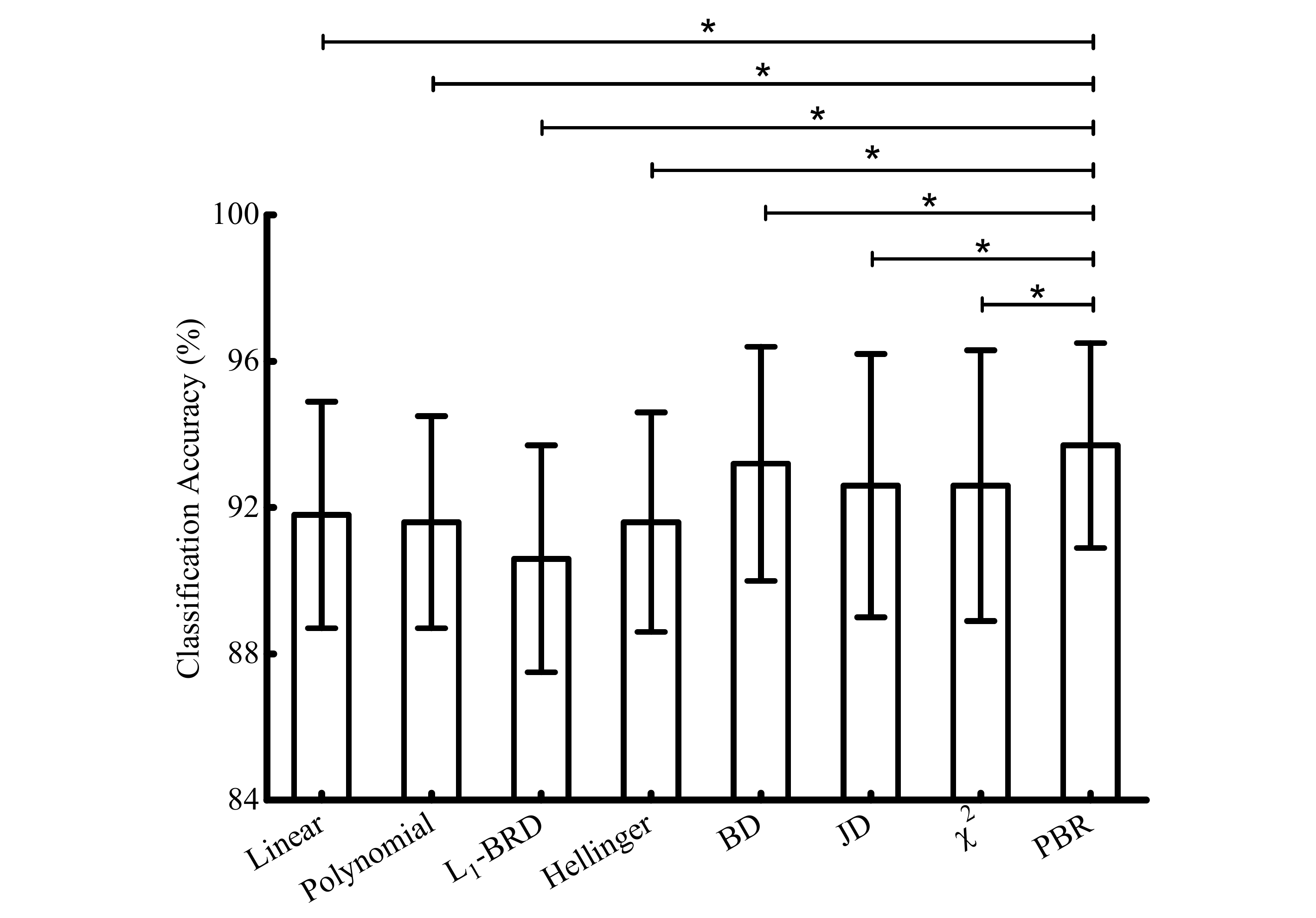}
	\end{subfigure}
	\begin{subfigure}[b]{0.49\textwidth}
		\includegraphics[width=\textwidth]{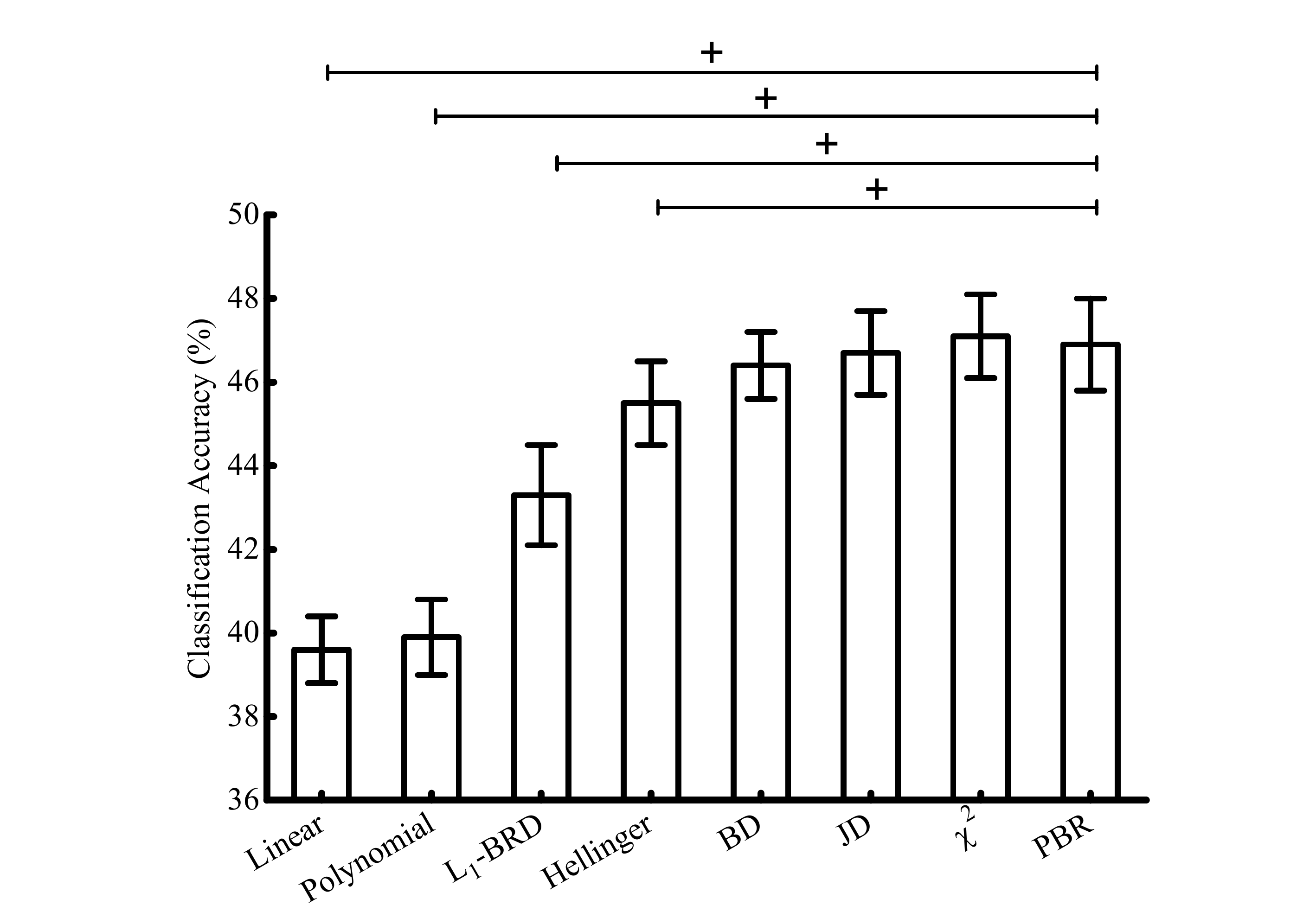}
	\end{subfigure}
\caption{Classification accuracy (percent) for the (a) LFW-Cat data sets using 25 training images per class and (b) Caltech 101 data set using 30 training images per class. Means and standard deviations are reported. PBR significantly (Bonferroni corrected Wilcoxon-signed rank test) outperforms at $\alpha$ = 0.05 (95\% confidence) indicated by `+' and $\alpha$ = 0.005 (99.5\% confidence) indicated by `$\ast$'.}
\label{fig:category}
\end{figure*}

The detection of cat heads and faces have attracted the recent interest of researchers, reflecting their popularity on the internet and as human companions. Although, sharing a similar face geometry to humans, approaches for detecting human faces can't be directly applied to cats because of the high intra-class variation among the facial features and textures of cats as compared to humans. This motivated us to test the PBR kernel on a cat vs human binary classification task. We used the LFW and cat data sets which consist of 13,233 human images and 9,997 cat images respectively. Sample images are shown in Fig.~\ref{fig:Human_cats}. We randomly selected 3000 images from each category to avoid a high computational workload. After resizing the images to have a maximum dimension of 250 pixels while maintaining the aspect ratio, we extracted the PRICoLBP feature using the 6 template configuration. 25 training images per class were used for training and the rest for testing.
\newline
\indent The Caltech-101 data set is an important benchmark data set for image classification. This contains 9,144 images under 102 categories (101 diverse classes and one background class). The number of images per class varies from 31 to 800. These images exhibit high intra-class variations and vary in dimensions. Hence we resized the images to have a maximum dimension of 256 pixels (while maintaining the aspect ratio). The protocol we used was standard; 30 images per class for training and up to 50 images per class for testing. We applied PRICoLBP's 6 template configuration which produced a 3,540 dimensional feature.
\newline
\indent Experimental results presented in Fig.~\ref{fig:category} show that PBR achieves significantly (99.5\% confidence) superior performance compared to all other methods for human vs cat binary classification. For the Caltech 101 data set, PBR was comparable to {$\chi^2$} distance, while significantly outperforming baseline kernels and distance metrics ({$L_1$-BRD} and Hellinger). Moreover, Table~\ref{tab:categorylevelcaltech} shows that PBR, in comparison to Caltech 101, compared with BD, JD and $\chi^2$ distance, yields the best performance in 51 categories out of all 102 categories.

\begin{sidewaystable}
	\footnotesize
  \caption{Comparison of category-wise accuracy (percent) for PBR, BD, JD and $\chi^2$ distance measures on the Caltech 101 data set. The best performance is indicated in boldface.}
	\label{tab:categorylevelcaltech}
	\centering
		\begin{tabular}{cccccccccccccccc}
		\hline
		Category & PBR & BD & JD & $\chi^2$ & Category & PBR & BD & JD & $\chi^2$ & Category & PBR & BD & JD & $\chi^2$
		\\
		\hline 
		car side & \textbf{100.0} & \textbf{100.0} & \textbf{100.0} & 99.8 & okapi & 56.7 & 61.1 & 60.0 & \textbf{66.7} & bronsto & 30.8 & \textbf{33.8} & 31.5 & 32.3\\
		leopards & \textbf{100.0} & \textbf{100.0} & \textbf{100.0} & \textbf{100.0} & waterlil & \textbf{55.7} & 48.6 & 51.4 & 44.3 & rhino & 30.7 & 33.4 & 32.4 & \textbf{35.9}\\
		pagoda & \textbf{99.4} & 98.8 & 98.8 & 98.8 & flamingo & \textbf{53.3} & 50.7 & 52.0 & 50.7 & menorah & \textbf{30.2} & 28.4 & 29.8 & 29.2 \\
		minaret & \textbf{99.3} & 98.9 & 99.1 & 99.1 & yin yang & \textbf{53.0} & 50.0 & \textbf{53.0} & 50.7 & ceiling & 30.0 & 27.6 & 30.0 & \textbf{31.2}\\
		accordian & \textbf{97.6} & 97.2 & 97.2 & 97.2 & saxoph & \textbf{52.0} & 47.0 & 48.0 & 49.0 & platypus & \textbf{30.0} & 25.0 & 22.5 & 25.0\\
		trilobite & \textbf{95.4} & 94.4 & 95.0 & 94.0 & watch & \textbf{51.8} & 46.8 & 49.2 & 49.2 & mayfly & 29.0 & 32.0 & 31.0 & \textbf{33.0}\\
		cellphone & \textbf{94.5} & 93.8 & 94.1 & 94.1 & stapler & 51.3 & 50.7 & \textbf{53.3} & 52.7 & cup & 28.5 & 30.7 & 28.9 & \textbf{32.2}\\
		motorbik & \textbf{94.4} & 93.0 & 94.2 & 93.4 & chandel & \textbf{50.4} & 49.0 & \textbf{50.4} & 48.2 & buddha & 28.0 & 30.2 & 28.8 & \textbf{30.8}\\
		faceseasy & \textbf{94.2} & 92.6 & 93.0 & 93.6 & wheelch & 49.0 & 53.8 & 51.4 & \textbf{57.2} & stegosa & 26.9 & 27.9 & \textbf{29.3} & 27.2\\
		faces & \textbf{92.2} & 91.4 & 91.6 & 91.4 & headpho & 48.3 & 47.5 & 47.5 & \textbf{49.2} & electric & 25.3 & 20.4 & 23.8 & \textbf{25.6}\\
		airplanes & \textbf{90.8} & 89.0 & 90.0 & 89.6 & strawber & \textbf{48.0} & 44.0 & 46.0 & 44.0 & gramoph & 25.2 & 26.2 & 25.7 & \textbf{28.6}\\
		dalmatian & \textbf{90.3} & 87.8 & 89.2 & 86.5 & ketch & \textbf{46.4} & 42.8 & 45.4 & 42.8 & rooster & 24.7 & 27.9 & 24.7 & \textbf{28.9}\\
		dollarbill & \textbf{88.6} & 85.5 & 86.8 & 87.7 & soccer & \textbf{46.2} & 43.8 & 45.6 & 45.9 & barrel & 24.7 & 23.5 & 23.5 & \textbf{25.9}\\
		stopsign & 87.6 & 86.5 & \textbf{88.5} & 87.9 & schooner & 45.5 & 46.7 & 45.8 & \textbf{47.9} & butterfly & \textbf{22.2} & 21.0 & 20.2 & 22.0\\
		joshua & \textbf{83.2} & 79.7 & 81.2 & 80.0 & pyramid & \textbf{44.8} & 43.3 & \textbf{44.8} & 44.1 & octopus & 22.0 & \textbf{26.0} & 22.0 & \textbf{26.0}\\
		ferry & \textbf{81.9} & 78.1 & \textbf{81.9} & 78.9 & tick & \textbf{44.2} & \textbf{44.2} & \textbf{44.2} & 43.7 & bass & 19.6 & 22.1 & 19.6 & \textbf{23.7}\\
		grandpi & \textbf{79.6} & 75.0 & 79.0 & 77.6 & panda & \textbf{43.8} & 42.5 & \textbf{43.8} & 42.5 & elephant & 18.8 & 21.8 & 18.5 & \textbf{22.1}\\
		snoopy & 76.0 & 72.0 & \textbf{78.0} & 74.0 & wrench & \textbf{42.2} & 37.8 & 41.1 & 41.1 & crocodile & \textbf{18.5} & 15.5 & 17.5 & 17.0\\
		windsor & \textbf{75.8} & 73.8 & 75.4 & 73.5 & bonsai & 41.2 & 43.6 & 41.8 & \textbf{44.0} & cougar & 18.2 & \textbf{24.1} & 22.4 & 22.4\\
		laptop & 71.6 & 72.8 & 73.4 & \textbf{73.8} & mandol & \textbf{40.8} & 39.2 & \textbf{40.8} & 40.0 & ewer & 17.4 & 18.2 & 16.4 & \textbf{19.2}\\
		scissors & 71.1 & 67.8 & \textbf{72.2} & 70.0 & hawksbi & \textbf{39.6} & 38.4 & 39.2 & 38.4 & pigeon & 17.3 & 14.0 & \textbf{18.0} & 14.7\\
		euphon & 67.9 & 67.9 & 68.8 & \textbf{69.1} & dolphin & 39.1 & 36.6 & \textbf{40.3} & 37.7 & chair & 14.7 & 14.1 & 14.4 & \textbf{16.3}\\
		pizza & 65.2 & 67.4 & 63.9 & \textbf{67.8} & helicopt & 38.2 & 35.8 & 36.0 & \textbf{40.4} & cannon & 13.8 & 18.5 & 15.4 & \textbf{20.8}\\
		garfield & \textbf{65.0} & \textbf{65.0} & \textbf{65.0} & \textbf{65.0} & lotus & \textbf{38.1} & 37.5 & 37.2 & 37.2 & beaver & 13.8 & 15.0 & 15.6 & \textbf{18.1}\\
		metron & \textbf{65.0} & 60.0 & \textbf{65.0} & \textbf{65.0} & croc.head & 37.6 & \textbf{41.9} & 38.1 & 39.5 & kangaroo & 13.0 & \textbf{15.8} & 13.6 & 15.6\\
		sunflow & \textbf{64.2} & 62.2 & 62.6 & 61.0 & nautilus & \textbf{36.0} & 31.2 & 35.2 & 34.8 & scorpion & 12.6 & 16.0 & 13.8 & \textbf{17.6}\\
		brain & 63.4 & \textbf{65.0} & 62.2 & 64.4 & wildcat & \textbf{35.0} & \textbf{35.0} & \textbf{35.0} & 30.0 & ibis & 12.0 & 15.8 & 13.4 & \textbf{16.0}\\
		binocular & \textbf{63.3} & \textbf{63.3} & \textbf{63.3} & \textbf{63.3} & starfish & 34.0 & 30.4 & \textbf{34.4} & 32.8 & dragonfly & 11.3 & 9.5 & \textbf{12.1} & 10.0 \\
		cougar & 62.8 & \textbf{65.9} & 64.1 & 65.6 & seahorse & \textbf{33.3} & 32.6 & 31.5 & 31.9 & lamp & 10.0 & \textbf{11.0} & \textbf{11.0} & 10.3 \\
		revolver & \textbf{60.4} & 54.6 & 59.4 & 57.2 & emu & 33.0 & 31.7 & 34.8 & \textbf{37.0} & crayfish & 8.8 & 9.2 & \textbf{10.3} & 8.5\\
		gerenuk & 60.0 & \textbf{75.0} & 55.0 & 72.5 & flamingo & \textbf{32.4} & 32.2 & \textbf{32.4} & 30.3 & lobster & \textbf{8.2} & 7.3 & \textbf{8.2} & \textbf{8.2}\\
		inline  & \textbf{60.0} & 50.0 & \textbf{60.0} & 50.0 & Ilama & 32.1 & 37.1 & 32.5 & \textbf{37.7} & anchor & 7.5 & \textbf{8.3} & \textbf{8.3} & \textbf{8.3}\\
		camera & 59.5 & 60.5 & 60.0 & \textbf{62.0} & umbrella & \textbf{31.3} & 30.0 & 31.1 & 28.7 & crab & 4.9 & \textbf{8.1} & 6.5 & 7.7 \\
		hedgehog & \textbf{58.3} & 55.4 & 57.1 & 52.9 & backgro & \textbf{30.8} & 28.4 & 30.0 & 27.4 & ant & 0.8 & 0.0 & 0.8 & \textbf{3.3}\\
		\hline 
		\end{tabular}
\end{sidewaystable}

\subsection{Evaluation of generalization capability}
Table~\ref{tab:support_vectors} summarizes the average number of SVs per chosen model for all methods evaluated in all data sets. A classifier which has a high number of SV's in proportion to training data can be said to have generalized poorly. By this token, we observe that PBR is consistently associated with the lowest number of SVs compared to the state-of-the-art distance measures. The only exception is IIT-II, where $\chi^2$ is associated with one less SV. Overall, this data indicates that PBR's improved performance is attributable to its superior generalization capability and not to overfitting.
\begin{sidewaystable}
	\caption{Comparison of the average number of SVs per chosen model for all methods evaluated in all data sets. The number given below the data set name denotes the number of training images.}
	\label{tab:support_vectors}
	\centering
		\begin{tabular}{cccccccccccccccccc}
		\cline{1-17} & \multicolumn{2}{c}{Brodatz} & \multicolumn{2}{c}{KTH} & \multicolumn{2}{c}{Kylberg} & \multicolumn{2}{c}{FMD} & \multicolumn{2}{c}{Swed Leaf} & IIT-I & IIT-II & USTB & MIT & Binary & Caltech\\
		\cline{2-17}
		Methods & 222 & 333 & 100 & 400 & 56 & 140 & 100 & 500 & 75 & 375 & 368 & 572 & 125 & 5,360 & 50 & 3,060\\
		\cline{1-17}
		Linear & 15 & 18 & 22 & 50 & 7 & 11 & 32 & 131 & 14 & 32 & 28 & 28 & 17 & 379 & 19 & 188\\
		Polynomial & 15 & 18 & 22 & 50 & 7 & 11 & 33 & 136 & 14 & 33 & 28 & 28 & 17 & 387 & 19 & 192\\
		BD & 19 & 26 & 31 & 76 & 11 & 13 & 38 & 168 & 22 & 50 & 59 & 57 & 32 & 499 & 30 & 273\\
		JD & 17 & 23 & 30 & 73 & 11 & 13 & 38 & 169 & 22 & 52 & 58 & 56 & 32 & 534 & 30 & 261\\
		{$\chi^2$} & 18 & 21 & 31 & 76 & 11 & 13 & 39 & 170 & 22 & 53 & 58 & 55 & 31 & 499 & 30 & 274\\
		{$L_1$-BRD} & 28 & 34 & 47 & 113 & 20 & 29 & 56 & 246 & 34 & 80 & 63 & 63 & 35 & 822 & 43 & 396\\
		Hellinger & 32 & 35 & 49 & 121 & 17 & 27 & 55 & 257 & 32 & 84 & 85 & 84 & 45 & 941 & 43 & 458\\
		PBR & 17 & 20 & 28 & 68 & 8 & 13 & 36 & 168 & 19 & 48 & 58 & 56 & 31 & 481 & 20 & 256\\
		\hline
		\end{tabular}
\end{sidewaystable}

\section{Conclusion}

In this paper, we challenged the identical distribution assumption fundamental to all existing distance measures. There were three parts to our work. First, we asked if this assumption is a true reflection of real-world data. Using statistical analysis, we found this not to be true. This result led to our second contribution, the construction of the PBR distance measure using the Poisson-Binomial distribution. This distance measure avoids the identical distribution assumption and accounts for non-identically distributed features. Further, unlike commonly used distance measures, PBR captures bin-to-bin dispersion. Finally, in the third part of our investigation, we evaluated PBR's performance in six different image classification/recognition applications using twelve benchmark data sets spanning a wide variety of challenges encountered in computer vision. Our experimental results demonstrated that PBR outperforms state-of-the-art distance measures for most data sets and achieves comparable performance on the rest. This outperformance is not explained by the semimetric nature of PBR because other compared semimetric distances did not perform as well. These results support the idea that distance measures which account for different distributions can improve performance in classification and recognition.

\section*{Acknowledgments}
The authors thank Temasek Life Sciences Laboratory for funding this work and Entopsis LLC (Miami, FL) for providing all computer hardware used in this study. The authors also thank statistician Joakim Ekstr\"om, PhD, for sparking our initial interest in Bernoulli distributions and Joshua Vogelstein for reviewing an early version of the manuscript.\\

\bibliographystyle{elsarticle-num}


\bibliography{References}

\begin{thebibliography}{10}
\expandafter\ifx\csname url\endcsname\relax
  \def\url#1{\texttt{#1}}\fi
\expandafter\ifx\csname urlprefix\endcsname\relax\def\urlprefix{URL }\fi
\expandafter\ifx\csname href\endcsname\relax
  \def\href#1#2{#2} \def\path#1{#1}\fi

\bibitem{Jacobs2000}
D.~W. Jacobs, D.~Weinshall, Y.~Gdalyahu, Classification with nonmetric
  distances: image retrieval and class representation, IEEE Transactions on
  Pattern Analysis and Machine Intelligence 22~(6) (2000) 583--600.

\bibitem{Sebe2000}
N.~Sebe, M.~S. Lew, D.~P. Huijsmans, Toward improved ranking metrics, IEEE
  Transactions on Pattern Analysis and Machine Intelligence 22~(10) (2000)
  1132--1143.

\bibitem{zhang2007}
J.~Zhang, M.~Marsza{\l}ek, S.~Lazebnik, C.~Schmid, {Local features and kernels
  for classification of texture and object categories: A comprehensive study},
  International Journal of Computer Vision 73~(2) (2007) 213--238.

\bibitem{chapelle1999}
O.~Chapelle, P.~Haffner, V.~N. Vapnik, Support vector machines for
  histogram-based image classification, IEEE transactions on Neural Networks
  10~(5) (1999) 1055--1064.

\bibitem{scheirer2014}
W.~J. Scheirer, M.~J. Wilber, M.~Eckmann, T.~E. Boult, Good recognition is
  non-metric, Pattern Recognition 47~(8) (2014) 2721--2731.

\bibitem{Tversky1982}
A.~Tversky, I.~Gati, Similarity, separability, and the triangle inequality,
  Psychological Review 89~(2) (1982) 123.

\bibitem{Tillman2009}
R.~E. Tillman, Structure learning with independent non-identically distributed
  data, in: Proceedings of the 26th Annual International Conference on Machine
  Learning, ACM, 2009, pp. 1041--1048.

\bibitem{khanna2006}
N.~Khanna, A.~K. Mikkilineni, A.~F. Martone, G.~N. Ali, G.~T.-C. Chiu, J.~P.
  Allebach, E.~J. Delp, A survey of forensic characterization methods for
  physical devices, Digital Investigation 3 (2006) 17--28.

\bibitem{yu2008}
J.~Yu, J.~Amores, N.~Sebe, P.~Radeva, Q.~Tian, Distance learning for similarity
  estimation, IEEE Transactions on Pattern Analysis and Machine Intelligence
  30~(3) (2008) 451--462.

\bibitem{li2015}
Y.~Li, S.~Wang, Q.~Tian, X.~Ding, {Feature representation for
  statistical-learning-based object detection: A review}, Pattern Recognition
  48~(11) (2015) 3542--3559.

\bibitem{luo2015}
Y.~Luo, T.~Liu, D.~Tao, C.~Xu, Multiview matrix completion for multilabel image
  classification, IEEE Transactions on Image Processing 24~(8) (2015)
  2355--2368.

\bibitem{zhu2016}
X.~Zhu, X.~Li, S.~Zhang, Block-row sparse multiview multilabel learning for
  image classification, IEEE transactions on cybernetics 46~(2) (2016)
  450--461.

\bibitem{krizhevsky2012}
A.~Krizhevsky, I.~Sutskever, G.~E. Hinton, Imagenet classification with deep
  convolutional neural networks, in: Advances in Neural Information Processing
  Systems, 2012, pp. 1097--1105.

\bibitem{russell2008}
B.~C. Russell, A.~Torralba, K.~P. Murphy, W.~T. Freeman, {LabelMe: a database
  and web-based tool for image annotation}, International Journal of Computer
  Vision 77~(1-3) (2008) 157--173.

\bibitem{ILSVRC15}
O.~Russakovsky, J.~Deng, H.~Su, J.~Krause, S.~Satheesh, S.~Ma, Z.~Huang,
  A.~Karpathy, A.~Khosla, M.~Bernstein, et~al., Imagenet large scale visual
  recognition challenge, International Journal of Computer Vision 115~(3)
  (2015) 211--252.

\bibitem{he2015}
K.~He, X.~Zhang, S.~Ren, J.~Sun, Deep residual learning for image recognition,
  arXiv preprint arXiv:1512.03385.

\bibitem{clevert2015}
D.-A. Clevert, T.~Unterthiner, S.~Hochreiter, {Fast and accurate deep network
  learning by exponential linear units (ELUs)}, arXiv preprint
  arXiv:1511.07289.

\bibitem{wang2016}
J.~Wang, Y.~Yang, J.~Mao, Z.~Huang, C.~Huang, W.~Xu, {CNN-RNN: A unified
  framework for multi-label image classification}, arXiv preprint
  arXiv:1604.04573.

\bibitem{tang2015}
K.~Tang, M.~Paluri, L.~Fei-Fei, R.~Fergus, L.~Bourdev, Improving image
  classification with location context, in: Proceedings of the IEEE
  International Conference on Computer Vision, 2015, pp. 1008--1016.

\bibitem{niu2012}
X.-X. Niu, C.~Y. Suen, {A novel hybrid CNN--SVM classifier for recognizing
  handwritten digits}, Pattern Recognition 45~(4) (2012) 1318--1325.

\bibitem{huang2006}
F.~J. Huang, Y.~LeCun, Large-scale learning with svm and convolutional for
  generic object categorization, in: 2006 IEEE Computer Society Conference on
  Computer Vision and Pattern Recognition (CVPR'06), Vol.~1, IEEE, 2006, pp.
  284--291.

\bibitem{kim2015}
S.~Kim, Z.~Yu, R.~M. Kil, M.~Lee, Deep learning of support vector machines with
  class probability output networks, Neural Networks 64 (2015) 19--28.

\bibitem{you2010}
C.~H. You, K.~A. Lee, H.~Li, {GMM-SVM kernel with a Bhattacharyya-based
  distance for speaker recognition}, IEEE Transactions on Audio, Speech, and
  Language Processing 18~(6) (2010) 1300--1312.

\bibitem{nguyen2010}
H.-G. Nguyen, R.~Fablet, J.-M. Boucher, Spatial statistics of visual keypoints
  for texture recognition, in: European Conference on Computer Vision,
  Springer, 2010, pp. 764--777.

\bibitem{guo2013}
J.~Guo, Z.~Qiu, C.~Gurrin, Exploring the optimal visual vocabulary sizes for
  semantic concept detection, in: Content-Based Multimedia Indexing (CBMI),
  2013 11th International Workshop on, IEEE, 2013, pp. 109--114.

\bibitem{Vempati2010}
S.~Vempati, A.~Vedaldi, A.~Zisserman, C.~Jawahar, {Generalized RBF feature maps
  for efficient detection}, in: Proceedings of the British Machine Vision
  Conference, 2010, pp. 1--11.

\bibitem{tran2011}
H.~D. Tran, H.~Li, {Probabilistic distance SVM with Hellinger-exponential
  kernel for sound event classification}, in: IEEE International Conference on
  Acoustics, Speech and Signal Processing (ICASSP), IEEE, 2011, pp. 2272--2275.

\bibitem{vedaldi2012}
A.~Vedaldi, A.~Zisserman, Efficient additive kernels via explicit feature maps,
  IEEE Transactions on Pattern Analysis and Machine Intelligence 34~(3) (2012)
  480--492.

\bibitem{Rubner2000}
Y.~Rubner, C.~Tomasi, L.~J. Guibas, {The earth mover's distance as a metric for
  image retrieval}, International Journal of Computer Vision 40~(2) (2000)
  99--121.

\bibitem{werman1985}
M.~Werman, S.~Peleg, A.~Rosenfeld, A distance metric for multidimensional
  histograms, Computer Vision, Graphics, and Image Processing 32~(3) (1985)
  328--336.

\bibitem{haasdonk2010}
B.~Haasdonk, E.~Pekalska, {Classification with kernel mahalanobis distance
  classifiers}, in: Advances in Data Analysis, Data Handling and Business
  Intelligence, Springer, 2010, pp. 351--361.

\bibitem{hu2014}
W.~Hu, N.~Xie, R.~Hu, H.~Ling, Q.~Chen, S.~Yan, S.~Maybank, Bin ratio-based
  histogram distances and their application to image classification, IEEE
  Transactions on Pattern Analysis and Machine Intelligence 36~(12) (2014)
  2338--2352.

\bibitem{le1960}
L.~Le~Cam, et~al., {An approximation theorem for the Poisson binomial
  distribution}, Pacific J. Math 10~(4) (1960) 1181--1197.

\bibitem{Shen2013}
H.~Shen, N.~Zamboni, M.~Heinonen, J.~Rousu, {Metabolite identification through
  machine learning—tackling CASMI challenge using FingerID}, Metabolites
  3~(2) (2013) 484--505.

\bibitem{Lai2012}
A.~C. Lai, A.~N.~N. Ba, A.~M. Moses, Predicting kinase substrates using
  conservation of local motif density, Bioinformatics 28~(7) (2012) 962--969.

\bibitem{Niida2012}
A.~Niida, S.~Imoto, T.~Shimamura, S.~Miyano, Statistical model-based testing to
  evaluate the recurrence of genomic aberrations, Bioinformatics 28~(12) (2012)
  i115--i120.

\bibitem{Cazier2012}
J.-B. Cazier, C.~C. Holmes, J.~Broxholme, {GREVE: Genomic Recurrent Event
  ViEwer to assist the identification of patterns across individual cancer
  samples}, Bioinformatics 28~(22) (2012) 2981--2982.

\bibitem{Zhou2010}
H.~Zhou, M.~E. Sehl, J.~S. Sinsheimer, K.~Lange, et~al., Association screening
  of common and rare genetic variants by penalized regression, Bioinformatics
  26~(19) (2010) 2375--2382.

\bibitem{Wilm2012}
A.~Wilm, P.~P.~K. Aw, D.~Bertrand, G.~H.~T. Yeo, S.~H. Ong, C.~H. Wong, C.~C.
  Khor, R.~Petric, M.~L. Hibberd, N.~Nagarajan, {LoFreq: a sequence-quality
  aware, ultra-sensitive variant caller for uncovering cell-population
  heterogeneity from high-throughput sequencing datasets}, Nucleic Acids
  Research (2012) gks918.

\bibitem{conover1998}
W.~Conover, {Practical Nonparametric Statistics}, Wiley, 1998.

\bibitem{chen2010}
Y.~Chen, N.~Yu, B.~Luo, X.-w. Chen, {iLike: integrating visual and textual
  features for vertical search}, in: Proceedings of the 18th ACM international
  conference on Multimedia, ACM, 2010, pp. 221--230.

\bibitem{cieslak2009}
D.~A. Cieslak, N.~V. Chawla, {A framework for monitoring classifiers’
  performance: when and why failure occurs?}, Knowledge and Information Systems
  18~(1) (2009) 83--108.

\bibitem{bassam2010}
F.~Bassam, One-pass algorithms for large and shifting data sets, University of
  Southampton, School of Electronics and Computer Science, Doctoral Thesis,
  144pp. Available at: http://eprints. soton. ac. uk/159173/1/Thesis. pdf.

\bibitem{sharan2014}
L.~Sharan, R.~Rosenholtz, E.~H. Adelson, Accuracy and speed of material
  categorization in real-world images, Journal of Vision 14~(9) (2014) 12--12.

\bibitem{kylberg2011}
G.~Kylberg, \href{http://www.cb.uu.se/~gustaf/texture/}{{The Kylberg Texture
  Dataset v. 1.0}}, External report (Blue series)~35, Centre for Image
  Analysis, Swedish University of Agricultural Sciences and Uppsala University,
  Uppsala, Sweden (September 2011).
\newline\urlprefix\url{http://www.cb.uu.se/~gustaf/texture/}

\bibitem{yuan2005}
L.~Yuan, Z.~Mu, Z.~Xu, Using ear biometrics for personal recognition, in:
  Advances in Biometric Person Authentication, Springer, 2005, pp. 221--228.

\bibitem{kumar2012}
A.~Kumar, C.~Wu, Automated human identification using ear imaging, Pattern
  Recognition 45~(3) (2012) 956--968.

\bibitem{jia2011}
Y.~Jia, T.~Darrell, Heavy-tailed distances for gradient based image
  descriptors, in: Advances in Neural Information Processing Systems, 2011, pp.
  397--405.

\bibitem{chakraborty2015}
S.~Chakraborty, {Generating discrete analogues of continuous probability
  distributions-A survey of methods and constructions}, Journal of Statistical
  Distributions and Applications 2~(1) (2015) 1.

\bibitem{hwang2012}
Y.~Hwang, J.-S. Kim, I.~S. Kweon, Difference-based image noise modeling using
  skellam distribution, IEEE Transactions on Pattern Analysis and Machine
  Intelligence 34~(7) (2012) 1329--1341.

\bibitem{wang2004}
J.~Wang, Q.~Chen, Y.~Chen, Rbf kernel based support vector machine with
  universal approximation and its application, in: International Symposium on
  Neural Networks, Springer, 2004, pp. 512--517.

\bibitem{Sjoberg2013}
M.~Sj{\"o}berg, M.~Koskela, S.~Ishikawa, J.~Laaksonen, Large-scale visual
  concept detection with explicit kernel maps and power mean svm, in:
  Proceedings of the 3rd ACM conference on International conference on
  multimedia retrieval, ACM, 2013, pp. 239--246.

\bibitem{sharan2013}
L.~Sharan, C.~Liu, R.~Rosenholtz, E.~H. Adelson, Recognizing materials using
  perceptually inspired features, International Journal of Computer Vision
  103~(3) (2013) 348--371.

\bibitem{xie2014}
L.~Xie, J.~Wang, B.~Guo, B.~Zhang, Q.~Tian, Orientational pyramid matching for
  recognizing indoor scenes, in: Proceedings of the IEEE Conference on Computer
  Vision and Pattern Recognition, 2014, pp. 3734--3741.

\bibitem{brodatz1966}
P.~Brodatz, Textures: A Photographic Album for Artists and Designers, Dover
  Pubns, 1966.

\bibitem{hayman2004}
E.~Hayman, B.~Caputo, M.~Fritz, J.-O. Eklundh, On the significance of
  real-world conditions for material classification, in: European Conference on
  Computer Vision, Springer, 2004, pp. 253--266.

\bibitem{soderkvist2001}
O.~J.~O. S{\"o}derkvist, {Computer vision classification of leaves from Swedish
  trees}, Master's thesis, Link\"oping University, SE-581 83 Link{\"o}ping,
  Sweden, liTH-ISY-EX-3132 (September 2001).

\bibitem{Quattoni2009}
A.~Quattoni, A.~Torralba, Recognizing indoor scenes, in: Computer Vision and
  Pattern Recognition, 2009. CVPR 2009. IEEE Conference on, IEEE, 2009, pp.
  413--420.

\bibitem{fei2007}
L.~Fei-Fei, R.~Fergus, P.~Perona, Learning generative visual models from few
  training examples: An incremental bayesian approach tested on 101 object
  categories, Computer Vision and Image Understanding 106~(1) (2007) 59--70.

\bibitem{Huang2007}
G.~B. Huang, M.~Ramesh, T.~Berg, E.~Learned-Miller, Labeled faces in the wild:
  A database for studying face recognition in unconstrained environments, Tech.
  rep., Technical Report 07-49, University of Massachusetts, Amherst (2007).

\bibitem{Zhang2008}
W.~Zhang, J.~Sun, X.~Tang, Cat head detection-how to effectively exploit shape
  and texture features, in: European Conference on Computer Vision, Springer,
  2008, pp. 802--816.

\bibitem{qi2014}
X.~Qi, R.~Xiao, C.-G. Li, Y.~Qiao, J.~Guo, X.~Tang, Pairwise rotation invariant
  co-occurrence local binary pattern, IEEE Transactions on Pattern Analysis and
  Machine Intelligence 36~(11) (2014) 2199--2213.

\bibitem{albatal2014}
R.~Albatal, S.~Little, {Empirical exploration of extreme SVM-RBF parameter
  values for visual object classification}, in: International Conference on
  Multimedia Modeling, Springer, 2014, pp. 299--306.

\bibitem{gama2004}
J.~Gama, Functional trees, Machine Learning 55~(3) (2004) 219--250.

\bibitem{nadeau2003}
C.~Nadeau, Y.~Bengio, Inference for the generalization error, Machine Learning
  52~(3) (2003) 239--281.

\bibitem{deselaers2010}
T.~Deselaers, G.~Heigold, H.~Ney, {Object classification by fusing SVMs and
  Gaussian mixtures}, Pattern Recognition 43~(7) (2010) 2476--2484.

\end{thebibliography}

\newpage
\noindent {\bf{Muthukaruppan Swaminathan}} received his BE degree in Biomedical Engineering from Anna University, India in 2010. Subsequently, he received his MSc degree in Biomedical Engineering from Nanyang Technological University, Singapore in 2012. He is a currently a PhD candidate at Temasek Life Science Laboratory, Singapore and working toward his PhD degree in computational biology at the National University of Singapore, Singapore. His main interests are in the fields of pattern recognition, machine learning and data mining. \\

\noindent{\bf{Pankaj Kumar Yadav}} received his BTech degree from Indian Institute of Technology (IIT), Guwahati in 2012. During his time at Temasek Life Sciences Laboratory in Singapore in 2012, he worked on methods using the Discrete Fourier Transform to evaluate probability mass functions. \\

\noindent{\bf{Obdulio Piloto}} did his graduate thesis work at Johns Hopkins University School of Medicine on identifying mechanisms of drug resistance and developing novel means of targeting aberrant stem cells. He did his postdoctoral studies at Stanford University School of Medicine on cancer therapeutics. Obdulio is currently the co-founder and CEO of Entopsis, a startup company developing a novel cloud-based platform for medical diagnosis. His research interests are focused on machine learning as applied to diagnostics. \\

\noindent{\bf{Tobias Sj\"oblom}} did his graduate thesis work on growth factor inhibitors in experimental models of cancer at the Ludwig Institute for Cancer Research, Uppsala Branch. He did postdoctoral studies at Johns Hopkins University in Baltimore where he performed the first exome-wide mutational analyses of breast and colorectal cancers. Tobias is currently Associate Professor at Uppsala University in Sweden. His research interests concern the discovery and understanding of somatic mutations that cause common cancers, as well as development of computational tools for clinical sequencing. \\

\noindent{\bf{Ian Cheong}} received his PhD and pursued postdoctoral studies at Johns Hopkins University, Baltimore. After a stint as Visiting Scientist in Uppsala University working on statistical models for improving cancer mutation detection, he developed an interest in the application of pattern recognition to medical classification problems. Ian is currently a Principal Investigator at Temasek Life Sciences Laboratory and Assistant Professor at the National University of Singapore. 
\end{document}